\documentclass[10pt,twocolumn,letterpaper]{article}

\usepackage{iccv}
\usepackage{times}
\usepackage{epsfig}
\usepackage{graphicx}
\usepackage{amsmath}
\usepackage{amssymb}

\usepackage[pagebackref=true,breaklinks=true,colorlinks,bookmarks=false]{hyperref}

\iccvfinalcopy 

\def\iccvPaperID{1592} 

\def\RGB{{X}^{RGB}}
\def\Labels{\hat{Y}}
\def\Target{S^T}
\def\Canonical{S^N}
\def\2Dspace{S^{2D}}
\def\coordinates2D{p^{2D}}
\def\coordinates3D{\boldsymbol{p}^{T}}
\def\affineA{\boldsymbol{a}^R}
\def\affineB{\boldsymbol{b}^R}
\def\perspectiveF{\boldsymbol{f}}
\def\perspectiveC{\boldsymbol{c}}
\def\feature2{\boldsymbol{X}_\2Dspace}

\usepackage{graphicx}
\usepackage{times}
\usepackage{epsfig}
\usepackage{graphicx}
\usepackage{amsmath}
\usepackage{amssymb}
\usepackage{xcolor}
\usepackage{subcaption}
\usepackage{tabularx}
\usepackage{rotating}
\usepackage{verbatim}
\usepackage{xcolor,colortbl}
\usepackage{etoolbox}
\usepackage[normalem]{ulem}
\usepackage{booktabs}
\usepackage{multirow}
\usepackage{lipsum}
\usepackage{stmaryrd}
\usepackage{stackengine}
\usepackage{makecell}
\usepackage{pifont}
\usepackage{cancel}
\usepackage{adjustbox}
\usepackage{dblfloatfix}
\graphicspath{{imgs/}}
\usepackage[accsupp]{axessibility} %

\hypersetup{colorlinks}

\definecolor{LightGrey}{rgb}{.9,.9,.9}
\definecolor{White}{rgb}{1.,0.,1.}
\definecolor{first}{rgb}{.8,.0,.0}
\definecolor{second}{rgb}{.0,.6,.0}
\definecolor{third}{rgb}{.0,.0,.8}
\newcolumntype{g}{>{\columncolor{White}}c}

\definecolor{ceiling}{RGB}{214,  38, 40}   %
\definecolor{floor}{RGB}{43, 160, 4}     %
\definecolor{wall}{RGB}{158, 216, 229}  %
\definecolor{window}{RGB}{114, 158, 206}  %
\definecolor{chair}{RGB}{204, 204, 91}   %
\definecolor{bed}{RGB}{255, 186, 119}  %
\definecolor{sofa}{RGB}{147, 102, 188}  %
\definecolor{table}{RGB}{30, 119, 181}   %
\definecolor{tvs}{RGB}{160, 188, 33}   %
\definecolor{furniture}{RGB}{255, 127, 12}  %
\definecolor{objects}{RGB}{196, 175, 214} %

\definecolor{car}{rgb}{0.39215686, 0.58823529, 0.96078431}
\definecolor{bicycle}{rgb}{0.39215686, 0.90196078, 0.96078431}
\definecolor{motorcycle}{rgb}{0.11764706, 0.23529412, 0.58823529}
\definecolor{truck}{rgb}{0.31372549, 0.11764706, 0.70588235}
\definecolor{other-vehicle}{rgb}{0.39215686, 0.31372549, 0.98039216}
\definecolor{person}{rgb}{1.        , 0.11764706, 0.11764706}
\definecolor{bicyclist}{rgb}{1.        , 0.15686275, 0.78431373}
\definecolor{motorcyclist}{rgb}{0.58823529, 0.11764706, 0.35294118}
\definecolor{road}{rgb}{1.        , 0.        , 1.        }
\definecolor{parking}{rgb}{1.        , 0.58823529, 1.        }
\definecolor{sidewalk}{rgb}{0.29411765, 0.        , 0.29411765}
\definecolor{other-ground}{rgb}{0.68627451, 0.        , 0.29411765}
\definecolor{building}{rgb}{1.        , 0.78431373, 0.        }
\definecolor{fence}{rgb}{1.        , 0.47058824, 0.19607843}
\definecolor{vegetation}{rgb}{0.        , 0.68627451, 0.        }
\definecolor{trunk}{rgb}{0.52941176, 0.23529412, 0.        }
\definecolor{terrain}{rgb}{0.58823529, 0.94117647, 0.31372549}
\definecolor{pole}{rgb}{1.        , 0.94117647, 0.58823529}
\definecolor{traffic-sign}{rgb}{1.        , 0.        , 0.    }   

\makeatletter
\newcommand{\car@semkitfreq}{3.92}
\newcommand{\bicycle@semkitfreq}{0.03}
\newcommand{\motorcycle@semkitfreq}{0.03}
\newcommand{\truck@semkitfreq}{0.16}
\newcommand{\othervehicle@semkitfreq}{0.20}
\newcommand{\person@semkitfreq}{0.07}
\newcommand{\bicyclist@semkitfreq}{0.07}
\newcommand{\motorcyclist@semkitfreq}{0.05}
\newcommand{\road@semkitfreq}{15.30}  %
\newcommand{\parking@semkitfreq}{1.12}
\newcommand{\sidewalk@semkitfreq}{11.13}  %
\newcommand{\otherground@semkitfreq}{0.56}
\newcommand{\building@semkitfreq}{14.1}  %
\newcommand{\fence@semkitfreq}{3.90}
\newcommand{\vegetation@semkitfreq}{39.3}  %
\newcommand{\trunk@semkitfreq}{0.51}
\newcommand{\terrain@semkitfreq}{9.17} %
\newcommand{\pole@semkitfreq}{0.29}
\newcommand{\trafficsign@semkitfreq}{0.08}
\newcommand{\semkitfreq}[1]{{\csname #1@semkitfreq\endcsname}}

\newcommand{\ceiling@nyufreq}{1.37}
\newcommand{\floor@nyufreq}{17.58}
\newcommand{\wall@nyufreq}{15.26}
\newcommand{\window@nyufreq}{1.99}
\newcommand{\chair@nyufreq}{3.01}
\newcommand{\bed@nyufreq}{7.08}
\newcommand{\sofa@nyufreq}{4.70}
\newcommand{\table@nyufreq}{4.31}
\newcommand{\tvs@nyufreq}{0.47}
\newcommand{\furniture@nyufreq}{30.04}
\newcommand{\objects@nyufreq}{14.19}
\newcommand{\nyufreq}[1]{{\csname #1@nyufreq\endcsname}}

\usepackage[capitalize]{cleveref}
\crefname{section}{Sec.}{Secs.}
\Crefname{section}{Section}{Sections}

\Crefname{table}{Table}{Tables}
\crefname{table}{Tab.}{Tabs.}

\crefname{equation}{Eq.}{Eqs.}
\Crefname{equation}{Equation}{Equations}

\begin{document}

\title{NDC-Scene: Boost Monocular 3D Semantic Scene Completion\\in Normalized Device Coordinates Space}

\author{Jiawei Yao$^1$\thanks{These authors contribute equally to this work.} \quad Chuming Li$^{2,4}$\footnotemark[1]\quad  Keqiang Sun$^3$\footnotemark[1] \quad Yingjie Cai$^3$\quad  Hao Li$^3$\quad \\ Wanli Ouyang$^4$\footnotemark[2]\quad Hongsheng Li$^{3,4,5}$\thanks{Corresponding authors.}\\
$^1$ University of Washington
$^2$ The University of Sydney\\
$^3$  CUHK-SenseTime Joint Laboratory
$^4$ Shanghai AI Laboratory
$^5$ CPII under InnoHK\\
{\tt\small jwyao@uw.edu, chli3951@uni.sydney.edu.au, wanli.ouyang@sydney.edu.au,} \\
{\tt\small\{kqsun@link, caiyingjie@link, haoli@link, hsli@ee\}.cuhk.edu.hk}
}
\maketitle

\begin{abstract}

Monocular 3D Semantic Scene Completion (SSC) has garnered significant attention in recent years due to its potential to predict complex semantics and geometry shapes from a single image, requiring no 3D inputs. In this paper, we identify several critical issues in current state-of-the-art methods, including the Feature Ambiguity of projected 2D features in the ray to the 3D space, the Pose Ambiguity of the 3D convolution, and the Computation Imbalance in the 3D convolution across different depth levels. To address these problems, we devise a novel Normalized Device Coordinates scene completion network (NDC-Scene) that directly extends the 2D feature map to a Normalized Device Coordinates (NDC) space, rather than to the world space directly, through progressive restoration of the dimension of depth with deconvolution operations. Experiment results demonstrate that transferring the majority of computation from the target 3D space to the proposed normalized device coordinates space benefits monocular SSC tasks. Additionally, we design a Depth-Adaptive Dual Decoder to simultaneously upsample and fuse the 2D and 3D feature maps, further improving overall performance. Our extensive experiments confirm that the proposed method consistently outperforms state-of-the-art methods on both outdoor SemanticKITTI and indoor NYUv2 datasets. Our code are available at \href{https://github.com/Jiawei-Yao0812/NDCScene}{https://github.com/Jiawei-Yao0812/NDCScene}.

\end{abstract}

\section{Introduction}
Semantic Scene Completion (SSC)~\cite{song2017semantic} is a crucial task in 3D scene understanding~\cite{wojek2011monocular}~\cite{jaritz2019multi}, with wide applications like virtual reality, embodied AI, and autonomous driving, \etc. Despite the growing body of research on this topic, the majority of existing SSC solutions~\cite{roldao20223d}~\cite{cai2021semantic}~\cite{chen20203d} depend on input RGB image and corresponding 3D inputs such as depth image, truncated signed distance function (TSDF)\footnote{TSDF is a representation to encode depth volume, where each voxel
stores the distance value to its closest surface and the sign indicates whether the voxel is in visible spaces.},~\etc, to forecast volumetric occupancy and the corresponding semantic labels. However, the reliance on these 3D data often entails the use of specialized and costly depth sensors, and thus limits the further application of the Semantic Scene Completion algorithms. Recently, there is growing interest in monocular 3D semantic scene completion~\cite{cao2022monoscene}, which aims to reconstruct a volumetric 3D scene from a single RGB image, thus eliminating the requirement of the additional 3D inputs.

\begin{figure}[t!]
    \centering
    \captionsetup{font=footnotesize}
    \includegraphics[width=0.5\textwidth]{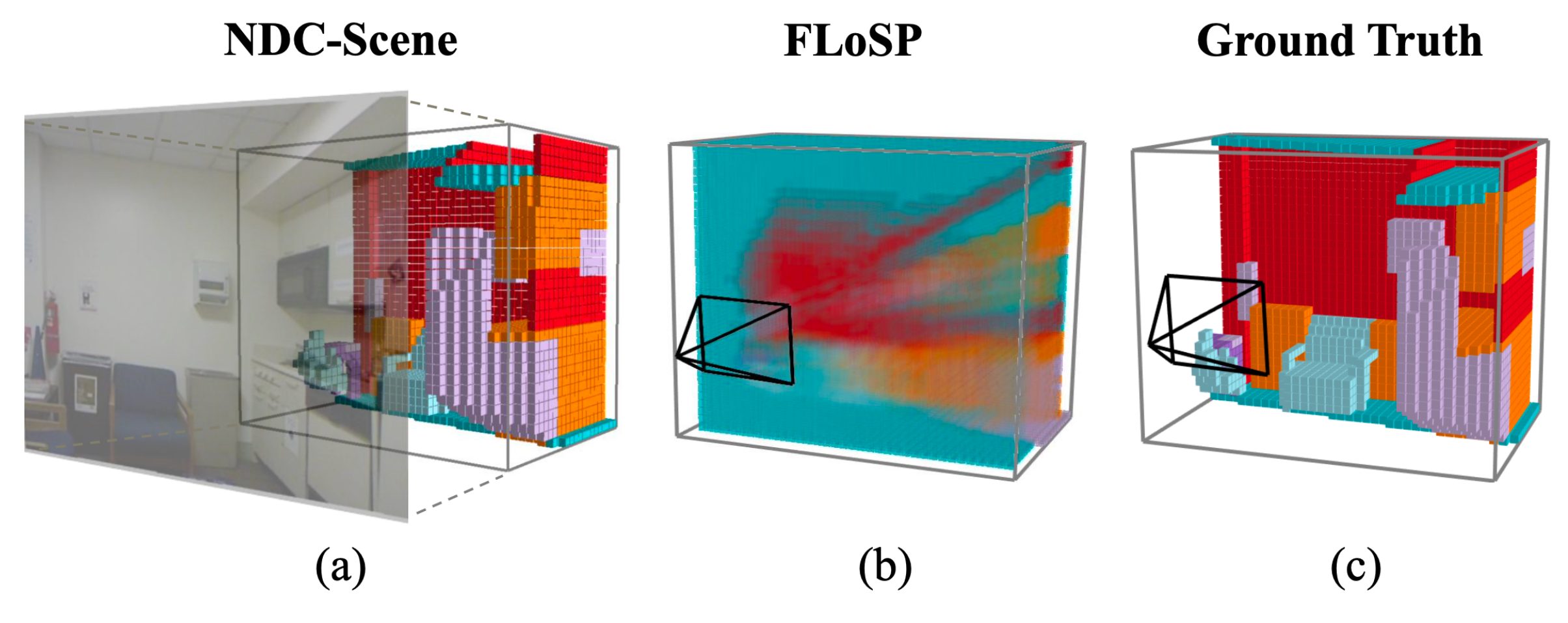}
    
    \caption{
    \textbf{Feature ambiguity}. We compare (a) the feature maps generated by the proposed dual decoder in the normalized device coordinates space, and (b) the feature maps projected through FLoSP~\cite{cao2022monoscene}, with reference to the ground truth demonstrated in (c). In (b), the multi-scale 2D features are projected along their line of sight, which introduces ambiguity in both feature size and feature depth. Conversely, in the normalized device coordinates space, the semantics and occupancy are implicitly restored, as exemplified in (a).}
    
    \label{fig:flosp}
\end{figure}


In the pioneering method Monoscene~\cite{cao2022monoscene}, the 2D features are lifted to the 3D space by inverting the perspective projection, where the same 2D features are propagated to different depths along the cast camera rays.
As depicted in Fig.~\ref{fig:flosp} (b), all the voxels on the projected position of the line sharing the same 2D feature at that specific position.
This approach broadcast the 2D feature in the 3D space, making it possible to employ a 3D UNet to predict the completed semantic scene volume.
 
However, we notice some ambiguities in the prior works, which can be summarized as Feature-Size Ambiguity (FSA), Feature-Depth Ambiguity (FDA) and Pose Ambiguity (PA).
The aforementioned projection gives rise to the FSA and FDA. With regard to FSA, the utilization of perspective projection results in the spread of 2D image features across a larger space as the depth increases. As shown in Fig.~\ref{fig:flosp} (b), this leads to variations in feature density across different depths. These 3D features with inconsistent density pose a challenge for the convolution kernel to discern the effective patterns. As for FDA, as shown in Fig.~\ref{fig:flosp} (b), each 2D feature pixel corresponds to a specific position and category, hence propagating 2D feature pixels to all the voxels along the ray makes the depth and category indistinguishable in the constructed 3D features.

\begin{figure}[!t]
    \centering
    \captionsetup{font=footnotesize}
    \includegraphics[width=0.48\textwidth]{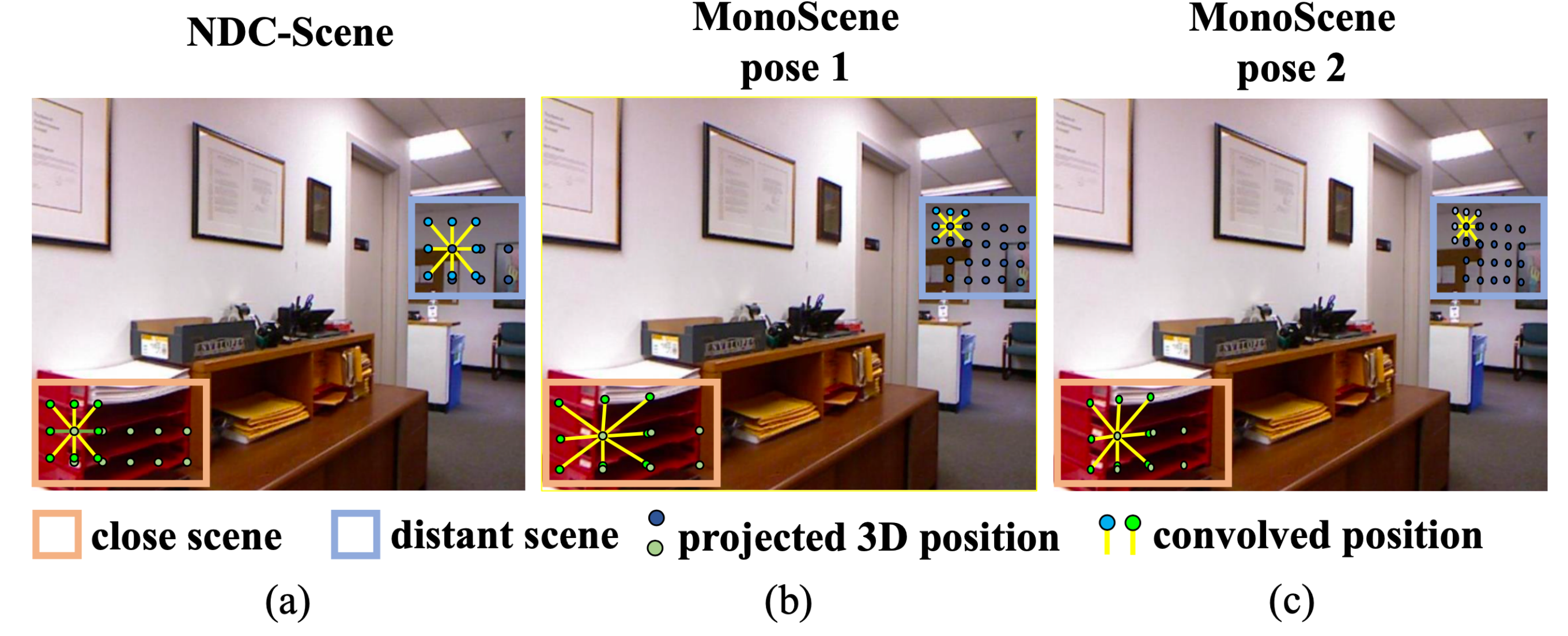}
    \caption{
    \textbf{Pose ambiguity and Imbalanced computation}. We illustrate the projected 2D positions from the 3D convolutions in (a) the proposed normalized device coordinates space and (bc) the target space. We chose two 3D convolution layer with the same stride and resolution, respectively in the two spaces. In (a), the projected positions uniformly distributes on the 2D pixels while in (bc) the  positions have an imbalanced allocation between the close and far scenes. Moreover, the convolution scope is not consistent among different choices of camera poses, especially manifesting in the different convolution offsets, as exemplified in (bc).}
    
    \label{fig:kernel}
\end{figure}

The Pose Ambiguity (PA) lies in the lack of camera extrinsic parameters. As illustrated in Fig.~\ref{fig:kernel} (b) and (c), given a certain input image, by supposing different camera extrinsic parameters, the relative positions between the convolved positions of a 3D convolution and the convolution center, when projected on the 2D feature map, should also transform accordingly. In other words, the 3D convolution should be conditioned on extrinsic parameters. However, prior works did not take the camera pose into account, which indicates the 3D convolution is performed based on an agnostic camera pose, entailing PA of the convolution scope. 

Besides, the perspective transformation between the target 3D space and the 2D camera plane introduces Computation Imbalance (CI) on the 2D feature map, which is also demonstrated in Fig.~\ref{fig:kernel} (b) and (c). Specifically, the convolved positions in the target 3D space, when projected on the 2D pixels, distributes quite sparse on the close scenes while dense on the far scenes. Such sparse computation allocation can hardly capture a comprehensive structural representation, from the rich details of structure or texture which usually exists in the 2D pixels projected from the close scenes, such as the red shelf in Fig.~\ref{fig:kernel} (b) and (c).

Based on the three ambiguities the computation imbalance noticed, we devise a novel framework named NDC-Scene.
Concretely, to alleviate FSA and FDA, the 3D feature maps are directly recovered in the NDC space, which strictly aligns with the image in hight and width, and extened in the depth-wise dimension. This methodology enables the implicit learning of precise occupancy and semantics among voxels, circumventing any erroneous inferences that may arise from 2D projection semantics. Furthermore, to address issues pertaining to PA and CI, we shift the majority computation units from the target 3D space to the NDC space. Extensive experiments on large-scale outdoor and indoor datasets demonstrate the superiority of our method to the existing state-of-the-art methods. The contributions can be summarized as follows:

\begin{figure*}[!th]
    \centering
    \captionsetup{font=footnotesize}
    \includegraphics[width=0.9\textwidth]{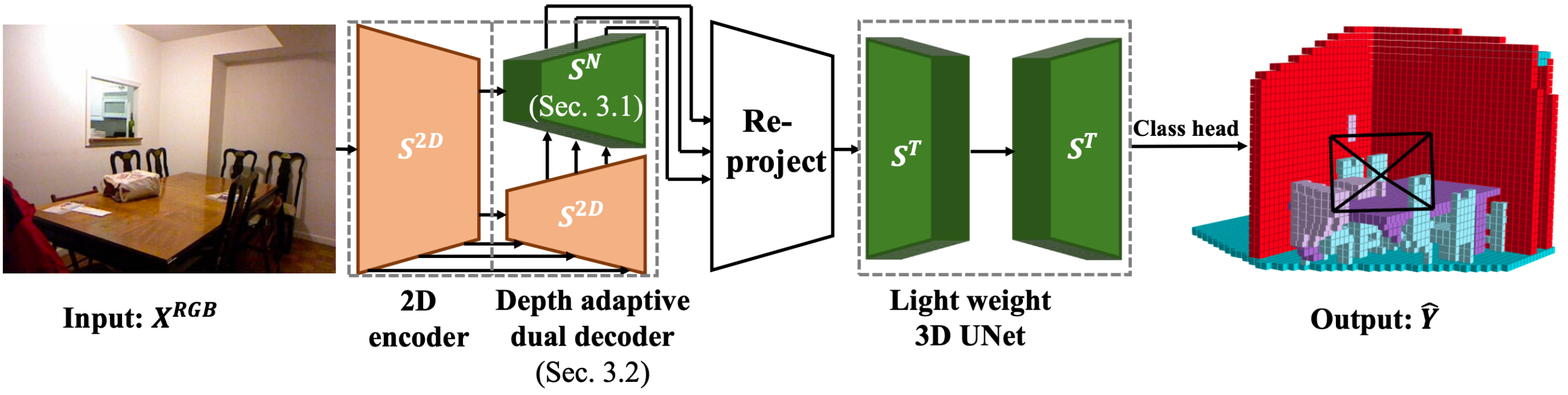}
    \caption{\textbf{NDC-Scene framework.} We first exploit an 2D image encoder to produce multi-scale 2D feature maps, followed by our Depth-Adaptive Dual Decoder (DADD Sec.~\ref{subsec:dual_decoder}) to restore the 3D feature map in $S^N$ (Sec.~\ref{subsec:canonical_space}), which is further re-projected to the target space $S^T$ to predict the SSC result via a light-weight 3D UNet and a class head.}
    \label{fig:framework}
    
\end{figure*}

\begin{itemize}
    \item According to the critical problems we noticed in the existing methods, we propose a novel method based on Normalized Device Coordinates (NDC) space, which is proved to be a better space to put the majority 3D computation units than the target 3D space.
    
    \item In conjunction with the aforementioned camera space prediction, a pioneering depth-adaptive dual decoder is introduced to jointly upsample both 3D and 2D features and integrate them, thereby attaining more resilient representations.
    
    \item Experimental results demonstrate that the proposed method outperforms state-of-the-art monocular semantic scene completion methods significantly on both outdoor and indoor datasets.
\end{itemize}

\section{Related Works}

\noindent\textbf{Single-View 3D Reconstruction} infers the object-level or scene-level 3D geometry from a single RGB image. Most existing works focus on the reconstruction of a single object, which exploits encoder-decoder structures to learn explicit~\cite{wu2016learning,girdhar2016learning,choy20163dr2n2,fan2017point,wang2018pixel2mesh,panos2018learning,liao2018deep,xie2020pix2vox,yu2020context,chen2020bspnet,piao2021inverting} 
or implicit~\cite{silberman2012indoor,mescheder2019occupancy,jeong2019deepsdf,peng2020convolutional,suncontrollable, sun2022cgof++} representations of 3D objects and reconstruct the object's volumetric or surface geometry.
A series of works~\cite{izadinia2017im2cad,gkioxari2019mesh,grabner2019location,zhang2020perceiving} extend this single-object 3D reconstruction to multi-object scenarios by reconstructing the instances detected in the image separately in a two-stage manner. For scene-level reconstruction, \cite{lee2017roomnet,huang2018holistic,sun2019horizonnet,nie2020total3dunderstanding} combine the estimations of overall layout and objects to obtain a sparse holistic 3D reconstruction of the scene. \cite{dahnert2021panoptic} lifts the features of 2D panoptic segmentation to 3D to obtain the dense estimation of indoor scenes. However, most existing methods still cannot perform dense reconstruction robustly in various type of scenes. Although recent work~\cite{cao2022monoscene} achieves dense semantic reconstruction of both indoor and outdoor scenes, it suffers from problems of PA and CI that limits its performance and robustness. In contrast, our method transfer most 3D computation units to a proposed normalized device coordinates space to avoid PA and CI, thus achieves better dense reconstruction in both indoor and outdoor scenes.

\vspace{0.5em}\noindent\textbf{3D Semantic Scene Completion} first defined in SSCNet~\cite{song2017semantic}, aims to jointly infer scene geometry and semantics given incomplete visual observations. Previous works~\cite{liu2018see,zhang2018efficient,li2019rgbd,zhang2019cascaded,li2020anisotropic,chen20203d,cai2021semantic} have extensively studied SSC for indoor small-scale scenes and achieved satisfactory results. With the emergence of large-scale outdoor scene datasets and demands in autonomous driving, a series of works~\cite{roldao2020lmscnet,yan2021sparse,rist2021semantic,cheng2021s3cnet} focus on the semantic completion of outdoor scenes, but such methods do not perform well in indoor scenes. At the same time, most existing works require RGB images along with additional geometric inputs, such as depth images~\cite{li2020anisotropic}, LiDAR point clouds~\cite{roldao2020lmscnet}, and Truncated Signed Distance Function (TSDF)~\cite{chen20203d,cai2021semantic}. VoxFormer~\cite{li2023voxformer} proposes to leverage a pretrained depth estimator as 3D prior. But the requirement of geometric data limit the application of these methods. A notable exception is MonoScene~\cite{cao2022monoscene}, which first investigates monocular SSC that rely only on a single-view RGB image as input for scene completion. MonoScene proposes the Features Line of Sight Projection (FLoSP) to bridge 2D and 3D features, achieving competitive performance with models with additional geometric inputs or 3D inputs and generalizes well to different types of scenes. Neverthless, the shared 2D features lifted to 3D rays via FLoSP results in both FSA and FDA, which limits its capacity in discerning the effective patterns of depth and density. 
 To improve the efficiency and performance, our method use a depth-adaptive dual decoder to restore the voxel feature on different depths in a more robust way, empowering it a strong representation of the occupancy and semantics among all depths.

\section{Methodology}
We consider a monocular 3D Semantic Scene Completion (SSC) task, which targets at predicting voxel-wise occupancy and semantics. Specifically, this task takes a single RGB image $\RGB$  as input and predicts volumetric labels $\Labels$ in a target 3D space $\Target$ with a shape $(H^T, W^T, D^T)$. The labels $\Labels$  $\in C^{H^{T}\times\:W^{T}\times\: 
D^{T}}$ are divided in $M+1$ categories $C=\left\{c_0, c_1, ..., c_M\right\}$, with $c_0$ denoting the free voxel and $\left\{c_1, ..., c_M\right\}$ being the semantic categories.

\paragraph{Overview} As mentioned above, current works in Monocular~\cite{cao2022monoscene} SSC exploit Features Line of Sight Projection(FLoSP) to project 2D features to the target 3D space $\Target$, thus introduces the ambiguity of both the size and the depth in the projected 3D feature. Further, the positions of the 3D convolution operation in $\Target$, when projected on the 2D feature map, suffers from the problems of pose ambiguity and imbalanced computation allocation. 

To tackle these problems, the proposed method extends the 2D feature map $\boldsymbol{X}^{2D}$ directly to the Normalized Device Coordinates space (NDC) $\Canonical$ via reconstructing the depth dimensional with deconvoluion operations (Sec.~\ref{subsec:canonical_space}). Such progressive reconstructed 3D features, compared with the shared 2D features projected along the camera ray in the FLoSP way, has the capacity of implicitly learning the density and depth of objects~\cite{alhashim2018high}, and thus relieves both FSA anf FDA.
Additionally, in $\Canonical$, the 2D projections of the 3D convolutions are evenly allocated. Such allocation has a stronger ability in capturing the structural representation of the rich details in the close scenes than the allocation of $\Target$.
 Finally, as $\Canonical$ is invariant to the camera pose, the 3D convolution kernel in $\Canonical$ has fixed offset, i.e., it has a consistent semantic representation among all camera poses and the pose ambiguity is avoided in $\Canonical$. 
 
 Besides, we design a depth adaptive dual decoder (Sec.~\ref{subsec:dual_decoder}) to simultaneously upsample 3D feature maps in $\Canonical$ and 2D feature maps derived from a 2D image encoder, respectively in two branches, as well as fuse them in each decoder layer to achieve more robust 3D semantic representations.
 
 The pipeline of the proposed method is plotted in Fig.~\ref{fig:framework}. At first, the input RGB image is encoded by a pre-trained 2D image encoder to generate multi-scale 2D feature maps. Afterwards, the proposed dual decoder is responsible reconstruct the 3D feature map in $\Canonical$, which is further re-projected into the target space $\Target$. Finally, a 3D UNet in $\Target$ processes the projected 3D features and predicts the completion result.
 In the proposed method, the 3D UNet in $\Target$ is quite light-weight, consisting of only downsample and upsample modules. While most 3D computation units is transferred to the 3D branch in the proposed dual decoder, which restore the 3D feature maps in $\Canonical$. We prove in Sec.~\ref{sec:experiment} that moving the majority of 3D computation cost from the target 3D space $\Target$ to the proposed $\Canonical$ brings obvious performance gains.

\begin{figure}[!t]
    \centering
    \captionsetup{font=footnotesize}
    \includegraphics[width=0.45\textwidth]{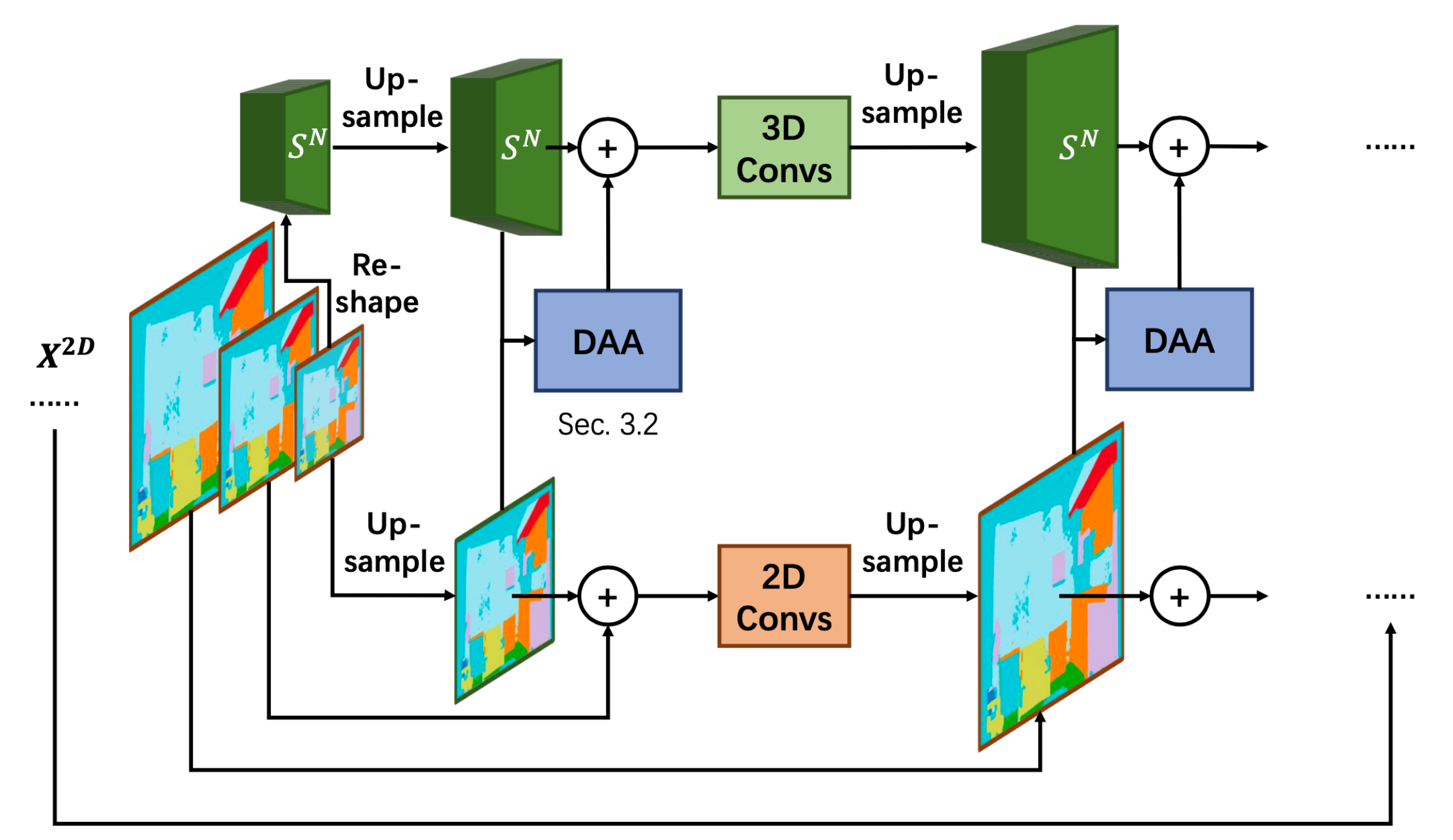}
    \caption{\textbf{Depth-adaptive dual decoder}. We infer the initial 3D feature map via a simple reshaping operation, and take the final feature map from the 2D image encoder as the the initial 2D feature map. In each decoder layer, the 2D  features and 3D features are first upsampled by a scale factor 2 to $\boldsymbol{X}_{s/2}^{2D}$ and $\boldsymbol{X}_{s/2}^{N}$. Then $\boldsymbol{X}_{s/2}^{2D}$ is fused into $\boldsymbol{X}_{s/2}^{N}$ via DAA (Sec.~\ref{subsec:dual_decoder}), and a residual 2D feature map is combined to $\boldsymbol{X}_{s/2}^{2D}$. Finally, $\boldsymbol{X}_{s/2}^{2D}$ and $\boldsymbol{X}_{s/2}^{N}$ are processed by 3D and 2D convolution units respectively.}
    
    \label{fig:dadd}
\end{figure}

\subsection{Normalized Device Coordinates Space}
\label{subsec:canonical_space}
\paragraph{Feature Ambiguity} Since the monocular SSC task~\cite{cao2022monoscene} assumes a RGB image from only a single view of point as input, it is impossible to back-project 2D features to their exact 3D correspondences due to lack of the guidance of depth. Current works~\cite{cao2022monoscene} exploits FLoSP to project 2D features to all possible locations in the target 3D space $\Target$ along their lines of sight. This practice, although shown to be effective, results in the ambiguity in the projected 3D features. This ambiguity can be categorized as Feature-Size Ambiguity (FSA) and Feature-Depth Ambiguity (FDA). As for FSA, the 2D image features are spread into larger space as the depth gets larger, as a result, such distribution with agnostic density makes it hard for the convolution kernel to determine the effective pattern. In a similar way, for depth, the 3D networks responsible for the SSC in $\Target$ can hardly distinguish the shared 2D features among all possible depths to discern the reasonable positions. An intuitive demonstration of feature ambiguity is shown in Fig.~\ref{fig:flosp} (b).

Also, we find two more imperceptible drawbacks existing in the mainstream monocular SSC works~\cite{cao2022monoscene}. To begin with, we first define the 2D space of the pixels in $\RGB$  as $\2Dspace$, and the coordinates in $\2Dspace$ as $\boldsymbol{p}^{2D}$, formally:
\begin{align}
    \2Dspace &= \left[0, W^{2D}\right] \times \left[0, H^{2D}\right], \\
    \boldsymbol{p}_{i,j}^{2D} &= \left(x_{i,j}^{2D}, y_{i,j}^{2D}\right) \in \2Dspace.
\end{align}
In common practice, a 2D feature map in the 2D space $\2Dspace$ is generated from a UNet structure, and then projected to the target 3D space, with the coordinates $\coordinates3D$ and space $\Target$ represented as:
\begin{align}
    S^{T} &= \left[0, W^{T}\right] \times \left[0, H^{T}\right] \times \left[0, D^{T}\right], \\
    \boldsymbol{p}_{i,j,k}^{T} &= \left(x_{i,j,k}^T, y_{i,j,k}^T, z_{i,j,k}^T\right) \in S^{T}.
\end{align}
The coordinates in $S^{T}$ and $\2Dspace$ are related via an affine transformation decided by the camera pose, formally, the extrinsic parameters $\affineA, \affineB$, followed by a perspective transformation, with parameters $\perspectiveF, \perspectiveC$, formally:
\begin{align}
    \left(x_{i,j,k}^R, y_{i,j,k}^R, d_{i,j,k}^R \right) &= \affineA \boldsymbol{p}_{i,j,k}^{T} + \affineB, \label{eq:rotate} \\
    \boldsymbol{p}_{i^\prime,j^\prime}^{2D} &= \perspectiveF\cdot\left(x_{i,j,k}^R, y_{i,j,k}^R\right) / d_{i,j,k}^R + \perspectiveC. \label{eq:prospective}
\end{align}
The 3D point $\boldsymbol{p}_{i,j,k}^{T}$ is first projected to the camera coordinate system via the affine transformation, with the projected coordinate $\left(x_{i,j,k}^R, y_{i,j,k}^R, d_{i,j,k}^R \right)$. This coordinate is further projected to $\boldsymbol{p}_{i^\prime,j^\prime}^{2D}$ in the 2D image space via the perspective transformation, with focal length $f$ and image center $c$. We find that both the two transformations introduce discrepancy between the convolution in the 3D space $\Target$ and the original pixel arrangement in $\2Dspace$.

\begin{figure}[!t]
    \captionsetup{font=footnotesize}
    \centering
    \setlength{\abovecaptionskip}{0.cm}
    \includegraphics[width=0.45\textwidth]{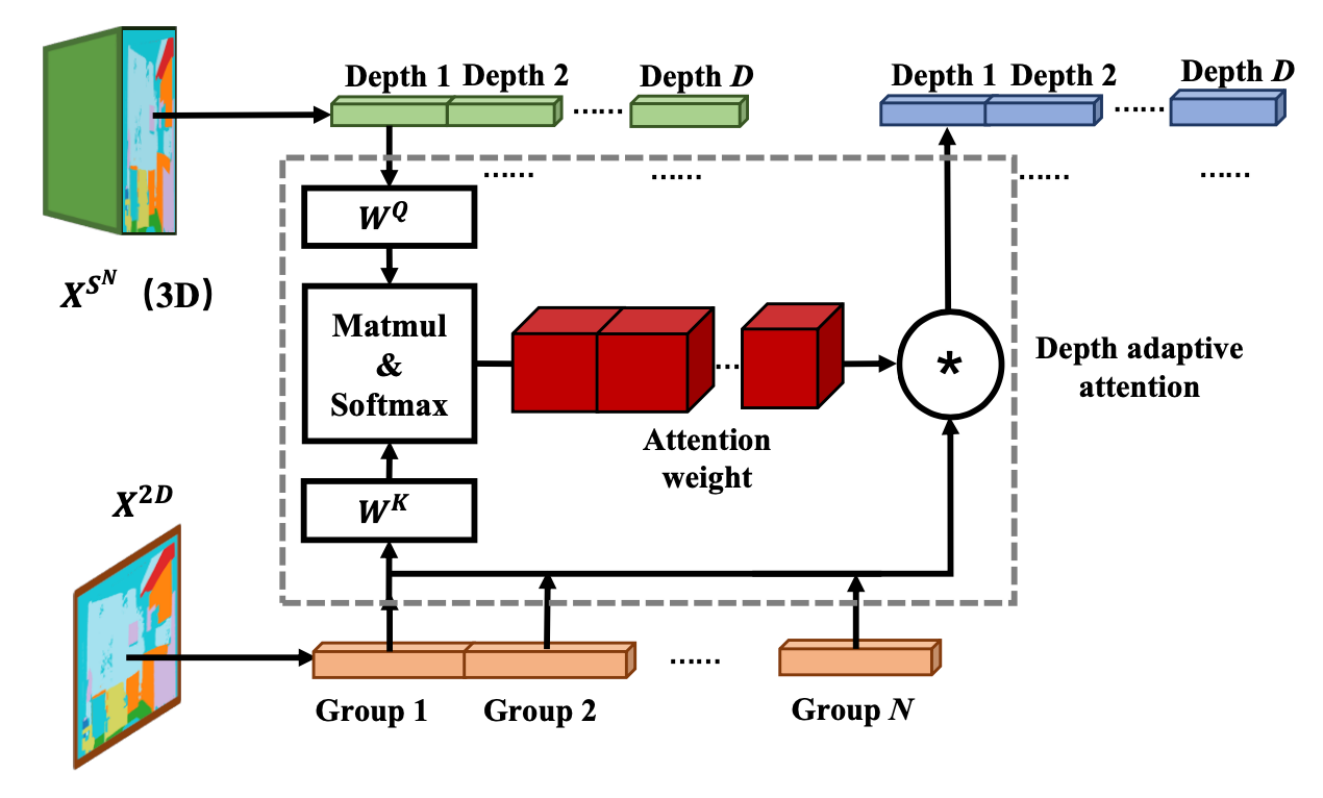}
    \caption{\textbf{Depth-adaptive attention}. We infer the attention matrix $\boldsymbol{a}$ via the inner-production between the 3D query feature and the 2D key feature on each group. We omit the value projection for computation reduction.} 
    
    \label{fig:daa}
\end{figure}

\paragraph{Pose Ambiguity} First, as a scene can be projected to multiple possible target 3D spaces for the SSC task, the affine transformation $\affineA, \affineB$ also has myriad possibilities. Currently, dataset providers usually selects $\affineA, \affineB$ following hand-crafted heuristic strategy, e.g., making a border of $\Target$ parallel to a border of the room, such as a wall. According to Eq.~\ref{eq:rotate}, the affine transformation decides the correspondence between the 2D coordinate $\boldsymbol{p}_{i,j}^{2D}$ and the 3D coordinate $\boldsymbol{p}_{i,j,k}^T$ in $\Target$, and thus the relative positions between the 2D projections of two 3D coordinates. However, the selection strategy of $\affineA, \affineB$ is agnostic to the SSC model, which means the convolution operation in $\Target$ is unaware of which positions in the 2D space it is really convolving on. To be specific, we consider a 3D convolution kernel $\boldsymbol{W}\in R^{(2K+1)^3 C_{in} C_{out}}$ which performs 3D convolution on a 3D position $\boldsymbol{p}_{i,j,k}^T$ in $\Target$,  with the kernel size $2K+1$. Its output on position $\boldsymbol{p}_{i, j, k}^T$ can be represented by:

\begin{align}
    \boldsymbol{O}_{i,j,k} &= \sum_{\left(i^\prime,j^\prime,k^\prime\right) \in I_r} \boldsymbol{W}_{i^\prime-i,j^\prime-j,k^\prime-k}\boldsymbol{X}_{i^\prime,j^\prime,k^\prime}^{T}, \\
    I_{i,j,k}^K &=\left\{i-K,...,i,...,i+K\right\} \notag \\
    &\times \left\{j-K,...,j,...,j+K\right\}  \notag \\
    &\times \left\{k-K,...,k,...,k+K\right\}. 
\end{align}
Where $I_{i,j,k}^K $ is the convolution scope in $\Target$ when the kernel convolves on  $\boldsymbol{p}_{i,j,k}^T$ and $X^{T} \in R^{C_{in}\times H^T\times W^T\times D^T}$ is the 3D feature map in $\Target$. Since the convolved 3D feature $\boldsymbol{X}_{i^\prime,j^\prime,k^\prime}^{T}$ is projected from the 2D feature map in $\2Dspace$, we further consider the back-projected 2D positions $\boldsymbol{p}_{I_{i,j,k}^K }^{2D}$ of the 3D positions $\boldsymbol{p}_{I_{i,j,k}^K }^{T}$ in this convolution scope, as well as the back-projected 2D position $\boldsymbol{p}_{i^\prime,j^\prime}^{2D}$ of the 3D position $\boldsymbol{p}_{i,j,k}^T$. As shown in Fig.~\ref{fig:kernel} (b) and (c), the relative position between $\boldsymbol{p}_{I_{i,j,k}^K }^{2D}$ and $\boldsymbol{p}_{i,j,k}^T$ is inconsistent between different selections of $\affineA, \affineB$, which implies that, given the same 2D feature map, 3D convolution in $\Target$ can hardly learns a robust semantic representation among different choice of $\affineA, \affineB$, named as Pose Ambiguity (PA).

\paragraph{Imbalanced Computation Allocation} Next, the perspective transformation $\perspectiveF, \perspectiveC$ introduces another problem between the 2D and 3D space, namely, the Imbalanced Computation Allocation. According to Eq.~\ref{eq:rotate}, $\boldsymbol{p}_{I_{i,j,k}^K }^{2D}$ distribute quite sparse on locations in close scenes due to the shallow depths, while very dense in the distant ones. Such imbalanced allocation is also shown in Fig.~\ref{fig:kernel} (b) and (c). However, the raw vision information are uniformly distributing among the 2D pixels, hence we argue that a uniform allocation of computation over the 2D space is more robust than the shown one. More intuitively, we notice that the close objects, when projected on the 2D image, contains more detailed structure or texture than the far ones. It means a sparse  computation allocation can hard capture a strong and comprehensive structural representation of the close objects. For example, in Fig.~\ref{fig:kernel} (b) and (c), the close scene contains rich detail of a red shelf, while the far scene are mostly composed of a simple wall.

\paragraph{Normalized Device Coordinates Space} To solve the three problems mentioned above, we propose a normalized device coordinates space $\Canonical$, with the coordinates $\boldsymbol{p}_{i,j,k}^{C}=\left(x_{i,j,k}, y_{i,j,k}, d_{i,j,k}\right)$. $\Canonical$ is derived via directly extending the location $\boldsymbol{p}_{i,j}^{2D}=\left(x_{i,j}, y_{i,j}\right)$ in the 2D pixels space with an additional dimension $d_{i,j,k}$, which is the depth to the camera plane. In this way, the 3D convolution operation has a consistent scope among different choices of affine transformations, as well as evenly distributed computation allocation among the 2D space $\2Dspace$, as shown in Fig.~\ref{fig:kernel} (a). It means that $\Canonical$ avoids the pose ambiguity and imbalanced computation allocation. Further, as described in Sec.~\ref{subsec:dual_decoder}, the progressively restored information in the dimension of depth endows the 3D feature map in $\Canonical$ a strong representation of the occupancy and semantics in different depths, especially compared to the shared feature among the sight of line in FLoSP. An intuitive comparison is provided in Fig.~\ref{fig:flosp} (a) and (b).


\begin{table*}
		\scriptsize
		\setlength{\tabcolsep}{0.0045\linewidth}
		\captionsetup{font = footnotesize}
		\newcommand{\nyuclassfreq}[1]{{~\tiny(\nyufreq{#1}\%)}}  %
		\centering
		\begin{tabular}{l|c|c|c c c c c c c c c c c|c}
			\toprule
			& & SC & \multicolumn{12}{c}{SSC} \\[-0.9em]
			Method
			& Input
			& {IoU}
			& \rotatebox{90}{\parbox{2cm}{\textcolor{ceiling}{$\blacksquare$} ceiling\nyuclassfreq{ceiling}}} 
			& \rotatebox{90}{\textcolor{floor}{$\blacksquare$} floor\nyuclassfreq{floor}}
			& \rotatebox{90}{\textcolor{wall}{$\blacksquare$} wall\nyuclassfreq{wall}} 
			& \rotatebox{90}{\textcolor{window}{$\blacksquare$} window\nyuclassfreq{window}} 
			& \rotatebox{90}{\textcolor{chair}{$\blacksquare$} chair\nyuclassfreq{chair}} 
			& \rotatebox{90}{\textcolor{bed}{$\blacksquare$} bed\nyuclassfreq{bed}} 
			& \rotatebox{90}{\textcolor{sofa}{$\blacksquare$} sofa\nyuclassfreq{sofa}} 
			& \rotatebox{90}{\textcolor{table}{$\blacksquare$} table\nyuclassfreq{table}} 
			& \rotatebox{90}{\textcolor{tvs}{$\blacksquare$} tvs\nyuclassfreq{tvs}} 
			& \rotatebox{90}{\textcolor{furniture}{$\blacksquare$} furniture\nyuclassfreq{furniture}} 
			& \rotatebox{90}{\textcolor{objects}{$\blacksquare$} objects\nyuclassfreq{objects}} 
			& mIoU\\
			\midrule

			LMSCNet$^\text{rgb}$~\cite{roldao2020lmscnet} & Occ & 33.93 & 4.49 & 88.41 & 4.63 & 0.25 & 3.94 & 32.03 & 15.44 & 6.57 & 0.02 & 14.51 & 4.39 & 15.88 
			\\
\hline
			AICNet$^\text{rgb}$~\cite{li2020anisotropic} & RGB \& Depth & 30.03 & 7.58 & 82.97 & 9.15 & 0.05 & 6.93 & 35.87 & 22.92 & 11.11 & 0.71 & 15.90 & 6.45 & 18.15 
			\\
\hline
			3DSketch$^\text{rgb}$~\cite{chen20203d}& RGB \& TSDF & 38.64 & 8.53 & 90.45&	9.94 & 5.67 & 10.64 & 42.29 & 29.21 & 13.88 & 9.38 & 23.83 & 8.19 & 22.91
			\\
\hline
            
			MonoScene~\cite{cao2022monoscene} & RGB & 	42.51 & 8.89 & 93.50 &  12.06 &  12.57 &  13.72 &  48.19 &  36.11 &  15.13 &  15.22 &  27.96 &  12.94 &  26.94\\
            NDC-Scene(ours) & RGB & \textcolor{blue}{\textbf{44.17}} & \textcolor{blue}{\textbf{12.02}} &	\textcolor{blue}{\textbf{93.51}} &	\textcolor{blue}{\textbf{13.11}} &	\textcolor{blue}{\textbf{13.77}}&	\textcolor{blue}{\textbf{15.83}}&	\textcolor{blue}{\textbf{49.57}}&	\textcolor{blue}{\textbf{39.87}}&	\textcolor{blue}{\textbf{17.17}}&	\textcolor{blue}{\textbf{24.57}}&	\textcolor{blue}{\textbf{31.00}}&	\textcolor{blue}{\textbf{14.96}}&	\textcolor{blue}{\textbf{29.03}} \\ 
			\bottomrule
		\end{tabular}
		\label{tab:nyu}
	
	\caption{\textbf{Quantitative comparison} against RGB-inferred baselines and the state-of-the-art monocular SSC method on NYUv2~\cite{silberman2012indoor}. The notations Occ, Depth and TSDF denote the occupancy grid(3D), depth map(2D) and TSDF array(3D), which are the 3D input required by the SSC baselines. For a fair comparison, all the three input are converted from the depth map predicted by a pretrained depth predictor ~\cite{bhat2021adabins}}

	\label{tab:NYUv2}
\end{table*}
\begin{figure*}
			
		\centering
		\footnotesize
        \captionsetup{font = footnotesize}
		\renewcommand{\arraystretch}{0.0}
		\setlength{\tabcolsep}{0.003\textwidth}
		\newcolumntype{P}[1]{>{\centering\arraybackslash}m{#1}}
		\captionsetup{font=scriptsize}
		\begin{tabular}{P{0.10\textwidth} P{0.115\textwidth} P{0.115\textwidth} P{0.115\textwidth} P{0.115\textwidth} P{0.115\textwidth} P{0.115\textwidth}}
			Input & 
   AICNet$^\text{rgb}$~\cite{li2020anisotropic} & 3DSketch$^\text{rgb}$~\cite{chen20203d} & MonoScene~\cite{cao2022monoscene} & NDC-Scene\scriptsize{(ours)} & Ground Truth
			\\[-0.1em]
			\includegraphics[width=\linewidth]{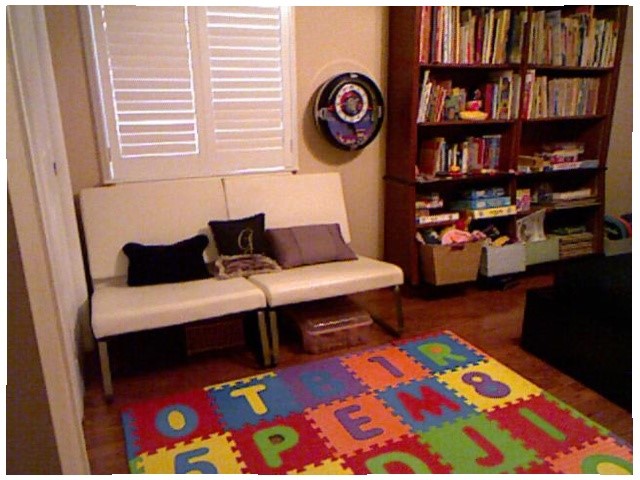} & 
			\includegraphics[width=\linewidth]{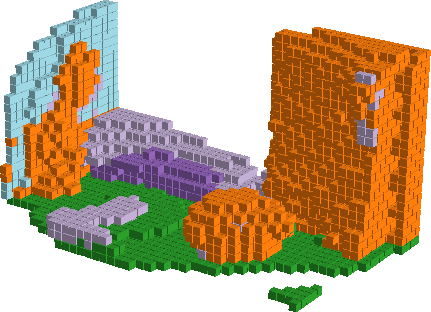} &
			\includegraphics[width=\linewidth]{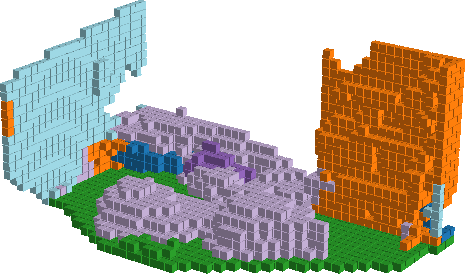} &
			\includegraphics[width=\linewidth]{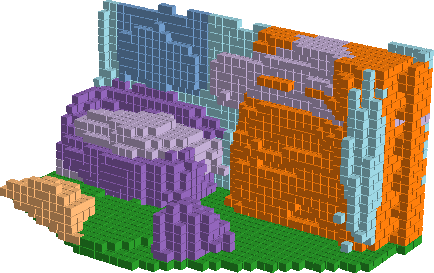} &
			\includegraphics[width=\linewidth]{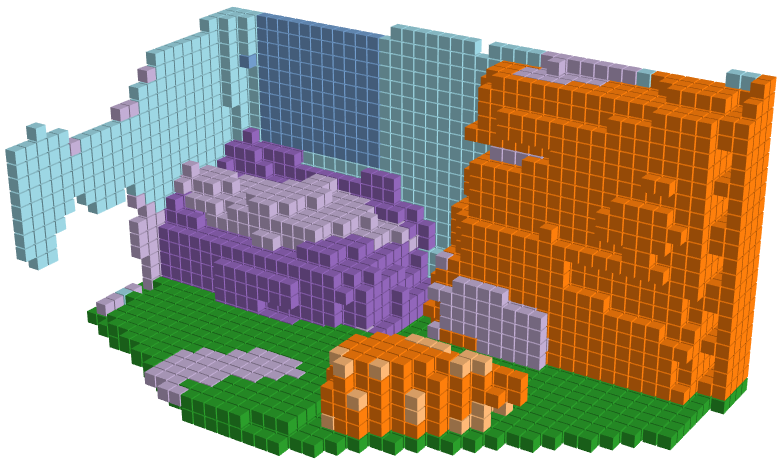} &
			\includegraphics[width=\linewidth]
            {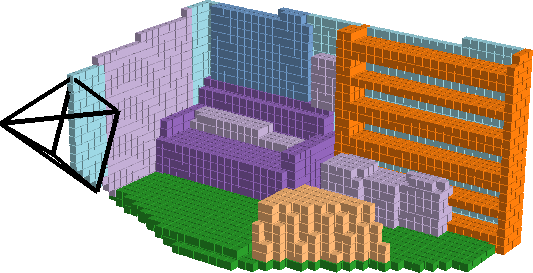} 
   
			\\[-0.3em]
			\includegraphics[width=\linewidth]{figure/nyu_qualitative/NYU0994_0000_2_color.jpg} & 
			\includegraphics[width=\linewidth]{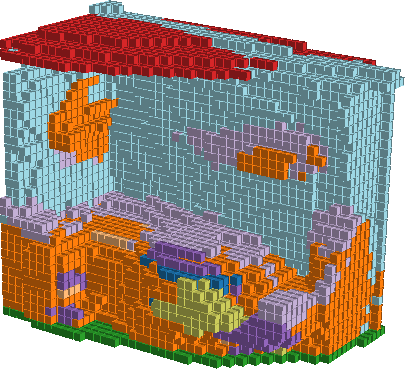} &
			\includegraphics[width=\linewidth]{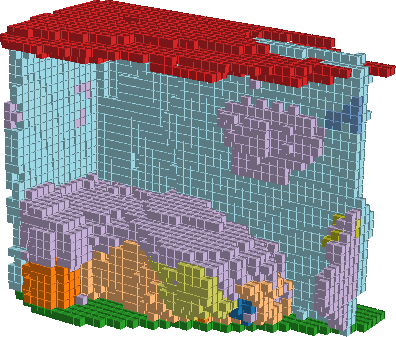} &
			\includegraphics[width=\linewidth]{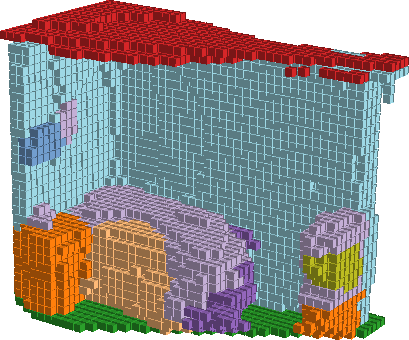} &
			\includegraphics[width=\linewidth]{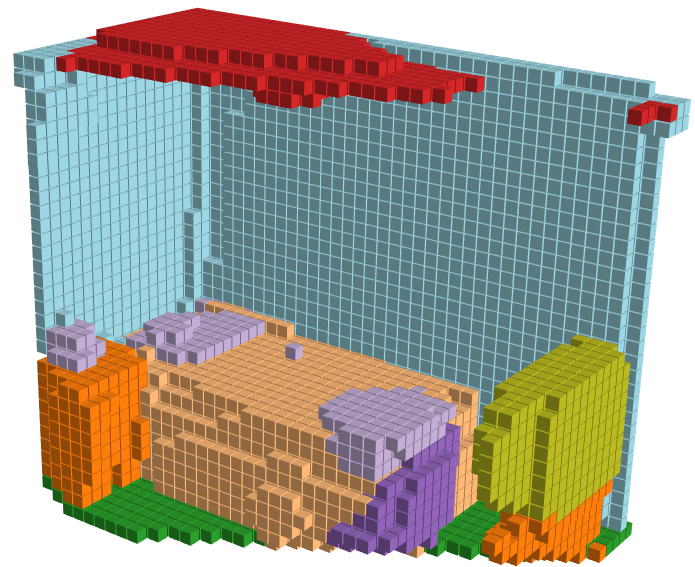} &
			\includegraphics[width=\linewidth]
            {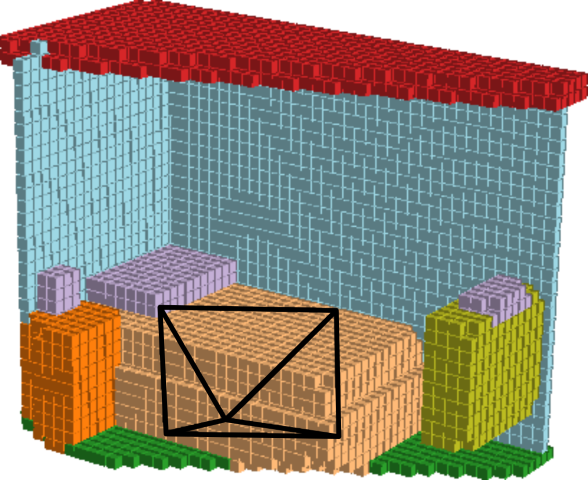}
			\\[-0.3em] 
			\includegraphics[width=\linewidth]{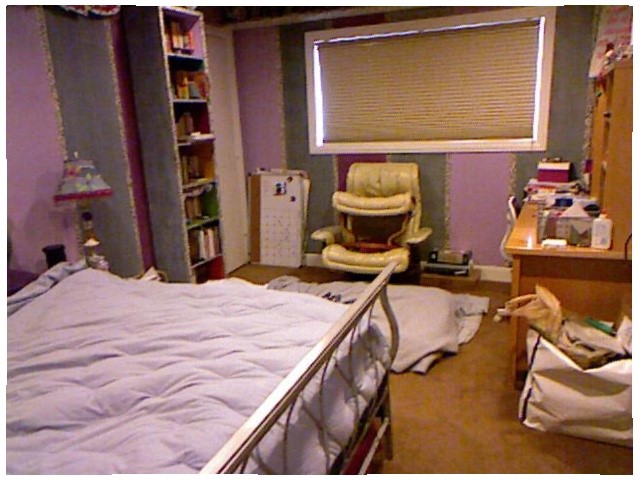} & 
			\includegraphics[width=\linewidth]{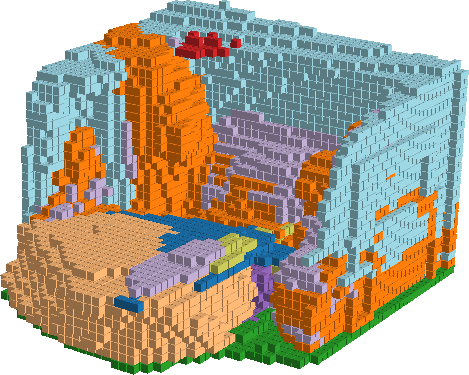} &
			\includegraphics[width=\linewidth]{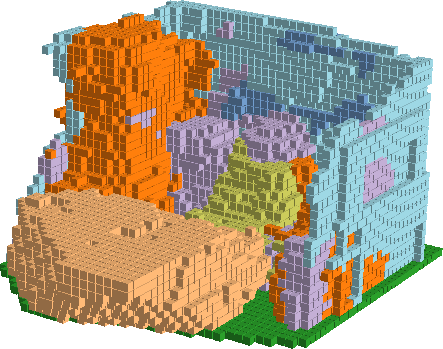} &
			\includegraphics[width=\linewidth]{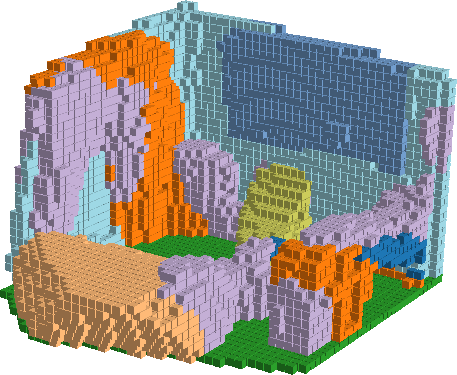} &
			\includegraphics[width=\linewidth]{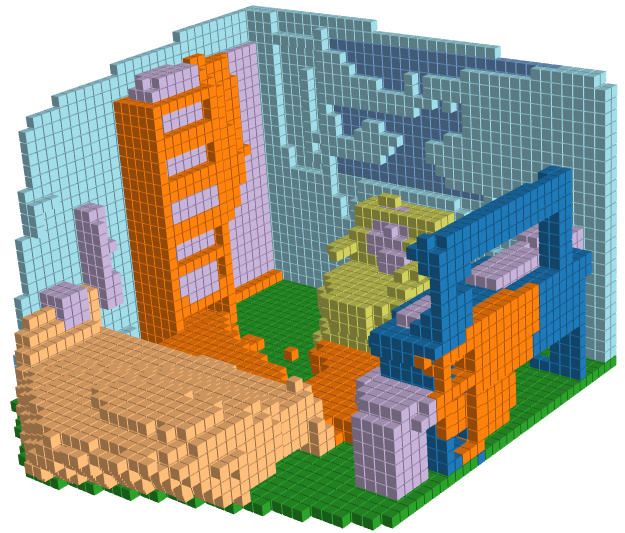} &
			\includegraphics[width=\linewidth]
            {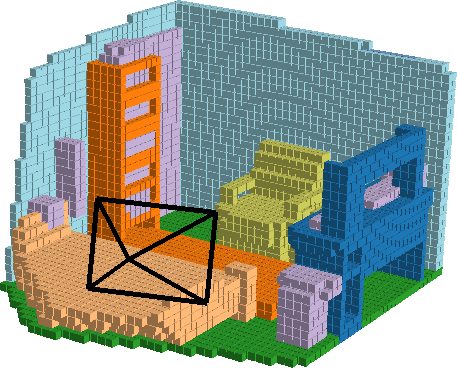}
			\\
			\multicolumn{6}{c}{
				\scriptsize
				\textcolor{ceiling}{$\blacksquare$}ceiling~
				\textcolor{floor}{$\blacksquare$}floor~
				\textcolor{wall}{$\blacksquare$}wall~
				\textcolor{window}{$\blacksquare$}window~
				\textcolor{chair}{$\blacksquare$}chair~
                    \textcolor{bed}{$\blacksquare$}bed~
				\textcolor{sofa}{$\blacksquare$}sofa~
				\textcolor{table}{$\blacksquare$}table~
				\textcolor{tvs}{$\blacksquare$}tvs~
				\textcolor{furniture}{$\blacksquare$}furniture~
				\textcolor{objects}{$\blacksquare$}objects
			}
		\end{tabular}

        \captionsetup{font = footnotesize}

	\caption{\textbf{Qualitative results on NYUv2~\cite{silberman2012indoor} (test set).} From left to right: (a) RGB input, (b) results of AICNet$^\text{rgb}$~\cite{li2020anisotropic}, (c) results of 3DSketch$^\text{rgb}$~\cite{chen20203d}, (d) results of  MonoScene~\cite{cao2022monoscene}, (e) ours results. NDC-Scene achieve higher voxel-level accuracy and better semantic predictions on NYUv2 (test set) compared with existing SSC baselines.}

	\label{fig:qualitative_NYU}
\end{figure*}

\subsection{Depth Adaptive Dual Decoder}
\label{subsec:dual_decoder}
We find that transferring the majority amount of 3D process from $\Target$ to the proposed normalized device coordinates space $\Canonical$ brings obvious performance gain, as illustrated in Sec.~\ref{sec:experiment}. To achieve a robust semantic representation in $\Canonical$, we propose a Depth Adaptive Dual Decoder (DADD). DADD simultaneously performs upsample on the 2D and 3D feature map, respectively in two branches of decoder layers, as well as fuse the 2D feature to 3D with a novel Depth Adaptive Attention (DAA) module.

\paragraph{Dual Decoder} The input of DADD is a 2D feature map $\feature2 \in R^{C_{in} \times W^{2D} \times H^{2D}}$ generated from a 2D encoder, which has a downscale stride $s$. Then, to achieve the initial 3D feature map, we exploit a reshaping operation to divide the dimension of channel into $D$ groups, each 
denoting an individual depth, formally:
\begin{align}
    \feature2 & \in R^{C_{in}\times W^{2D}\times H^{2D}} \notag \\
    & \stackrel{reshape}{\longrightarrow} \notag \\
    \boldsymbol{X}_s^{3D} & \in R^{C_{in}/D \times \ W^{2D}\times H^{2D} \times D}.
\end{align}
Afterwards, in each decoder layer, we first transform the 2D feature map $\feature2$ to a larger resolution $\boldsymbol{X}_{s/2}^{2D}$ with scale factor 2, following the common practice \cite{ronneberger2015u, cao2022monoscene} in the decoder of the 2D UNet, which upsamples $\feature2$ and add it to the residual feature map with the same resolution generated in the corresponding 2D encoder layer, followed by several 2D convolution units. Then, the 3D feature map $\boldsymbol{X}_s^{3D}$ is also upsampled to $\boldsymbol{X}_{s/2}^{3D}$. Afterwards, the upscaled $\feature2$ is fused into $\boldsymbol{X}_{s/2}^{3D}$ via the proposed DAA module, followed by several 3D convolution operations. We demonstrate more intuitive details in Fig.~\ref{fig:dadd}.

\paragraph{Depth Adaptive Attention} We assume that, with sufficiently large receptive field, a 2D feature at a 2D position $\boldsymbol{p}_{i,j}^{2D}$ in $\boldsymbol{X}_S^{2D}$ can aggregate the context information to implicitly infers both the surface and behind scenes at $\boldsymbol{p}_{i,j}^{2D}$. As fully verified in objective detection~\cite{chu2020detection,lin2017focal}, objects in different depth projected in the same 2D location, can be encoded in different channels of the detection head. In this paper, we also assume that the 3D semantic scene in different depth projected at $\boldsymbol{p}_{i,j}^{2D}$ can be encoded in different channel groups of the 2D feature $\boldsymbol{X}_{i,j}^{2D}$.
\begin{table*}
		\scriptsize
		\captionsetup{font=footnotesize}
		\setlength{\tabcolsep}{0.004\linewidth}
		\newcommand{\classfreq}[1]{{~\tiny(\semkitfreq{#1}\%)}}  %
		\centering
		\begin{tabular}{l|c|c|c c c c c c c c c c c c c c c c c c c|c}
			\toprule
			& & SC & \multicolumn{20}{c}{SSC} \\
			Method
			& SSC Input
			& {IoU}
			& \rotatebox{90}{\textcolor{road}{$\blacksquare$} road\classfreq{road}} 
			& \rotatebox{90}{\textcolor{sidewalk}{$\blacksquare$} sidewalk\classfreq{sidewalk}}
			& \rotatebox{90}{\textcolor{parking}{$\blacksquare$} parking\classfreq{parking}} 
			& \rotatebox{90}{\textcolor{other-ground}{$\blacksquare$} other-grnd\classfreq{otherground}} 
			& \rotatebox{90}{\textcolor{building}{$\blacksquare$} building\classfreq{building}} 
			& \rotatebox{90}{\textcolor{car}{$\blacksquare$} car\classfreq{car}} 
			& \rotatebox{90}{\textcolor{truck}{$\blacksquare$} truck\classfreq{truck}} 
			& \rotatebox{90}{\textcolor{bicycle}{$\blacksquare$} bicycle\classfreq{bicycle}} 
			& \rotatebox{90}{\textcolor{motorcycle}{$\blacksquare$} motorcycle\classfreq{motorcycle}} 
			& \rotatebox{90}{\textcolor{other-vehicle}{$\blacksquare$} other-veh.\classfreq{othervehicle}} 
			& \rotatebox{90}{\textcolor{vegetation}{$\blacksquare$} vegetation\classfreq{vegetation}} 
			& \rotatebox{90}{\textcolor{trunk}{$\blacksquare$} trunk\classfreq{trunk}} 
			& \rotatebox{90}{\textcolor{terrain}{$\blacksquare$} terrain\classfreq{ &terrain}} 
			& \rotatebox{90}{\textcolor{person}{$\blacksquare$} person\classfreq{person}} 
			& \rotatebox{90}{\textcolor{bicyclist}{$\blacksquare$} bicyclist\classfreq{bicyclist}} 
			& \rotatebox{90}{\textcolor{motorcyclist}{$\blacksquare$} motorcyclist.\classfreq{motorcyclist}} 
			& \rotatebox{90}{\textcolor{fence}{$\blacksquare$} fence\classfreq{fence}} 
			& \rotatebox{90}{\textcolor{pole}{$\blacksquare$} pole\classfreq{pole}} 
			& \rotatebox{90}{\textcolor{traffic-sign}{$\blacksquare$} traf.-sign\classfreq{trafficsign}} 
			& mIoU\\
			\midrule
	
            LMSCNet$^\text{rgb}$~\cite{roldao2020lmscnet} & Occ & 28.61 & 40.68 & 18.22 & 4.38 & 0.00 & 10.31 & 18.33 & 0.00 & 0.00 & 0.00 & 0.00 & 13.66 & 0.02 & 20.54 & 0.00 & 0.00 & 0.00 & 1.21 & 0.00 & 0.00 & 6.70 \\
\hline
            3DSketch$^\text{rgb}$~\cite{chen20203d} & RGB \& TSDF & 33.30 & 41.32 & 21.63 & 0.00 & 0.00 & 14.81 & 18.59 & 0.00 & 0.00 & 0.00 & 0.00 & 19.09 & 0.00 & 26.40 & 0.00 & 0.00 & 0.00 & 0.73 & 0.00 & 0.00 & 7.50  \\
            \hline

            AICNet$^\text{rgb}$~\cite{li2020anisotropic} & RGB \& Depth & 29.59 & 43.55 & 20.55 & 11.97 & 0.07 & 12.94 & 14.71 & 4.53 & 0.00 & 0.00 & 0.00 & 15.37 & 2.90 & 28.71 & 0.00 & 0.00 & 0.00 & 2.52 & 0.06 & 0.00 & 8.31  \\
            

            \hline
			MonoScene~\cite{cao2022monoscene} & RGB &  37.12 & 57.47 & 27.05 & 15.72 & 0.87 & 14.24 & 23.55 & 7.83 & 0.20 & 0.77 & 3.59 & 18.12 & 2.57 & 30.76 & 1.79 & 1.03 & 0.00 & 6.39 & 4.11 & 2.48 & 11.50 \\

            NDC-Scene(ours) & RGB & \textcolor{blue}{\textbf{37.24}} & \textcolor{blue}{\textbf{59.20}} & \textcolor{blue}{\textbf{28.24}} & \textcolor{blue}{\textbf{21.42}} & \textcolor{blue}{\textbf{1.67}} & \textcolor{blue}{\textbf{14.94}} & \textcolor{blue}{\textbf{26.26}} &  \textcolor{blue}{\textbf{14.75}} & \textcolor{blue}{\textbf{1.67}} & \textcolor{blue}{\textbf{2.37}} & \textcolor{blue}{\textbf{7.73}} & \textcolor{blue}{\textbf{19.09}} & \textcolor{blue}{\textbf{3.51}} & \textcolor{blue}{\textbf{31.04}} & \textcolor{blue}{\textbf{3.60}} & \textcolor{blue}{\textbf{2.74}} & 
            0.00 & \textcolor{blue}{\textbf{6.65}} & \textcolor{blue}{\textbf{4.53}} & \textcolor{blue}{\textbf{2.73}} & \textcolor{blue}{\textbf{12.70}}\\ 
			\bottomrule
		\end{tabular}\\
		\label{tab:semkitti}
 
	\caption{\textbf{Quantitative comparison} against RGB-inferred baselines and the state-of-the-art monocular SSC method on SemanticKITTI~\cite{behley2019semantickitti}. The notations Occ, Depth and TSDF denote the occupancy grid(3D), depth map(2D) and TSDF array(3D), which are the 3D input required by the SSC baselines. For a fair comparison, all the three input are converted from the depth map predicted by a pretrained depth predictor ~\cite{bhat2021adabins}}
	\label{tab:SemanticKITTI}
\end{table*}
\begin{figure*}	
		\centering
		\newcolumntype{P}[1]{>{\centering\arraybackslash}m{#1}}
		\setlength{\tabcolsep}{0.001\textwidth}
		\renewcommand{\arraystretch}{0.8}
		\footnotesize
		\captionsetup{font=scriptsize}
		\begin{tabular}{P{0.137\textwidth} P{0.137\textwidth} P{0.137\textwidth} P{0.137\textwidth} P{0.137\textwidth} P{0.137\textwidth}P{0.137\textwidth}}		
			Input & 
   AICNet$^\text{rgb}$~\cite{li2020anisotropic}  & LMSCNet$^\text{rgb}$~\cite{roldao2020lmscnet}  & MonoScene~\cite{cao2022monoscene} & NDC-Scene~\scriptsize{(ours)} & Ground Truth
			\\[-0.1em]
			\includegraphics[width=\linewidth]{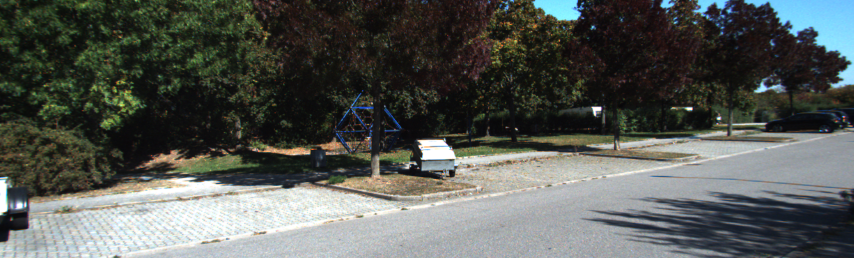} &  
			\includegraphics[width=\linewidth]
            {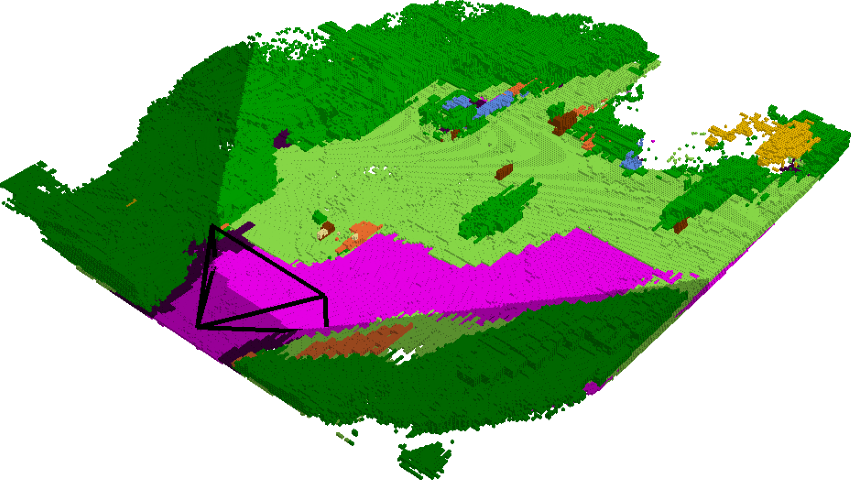} & 
			\includegraphics[width=\linewidth]{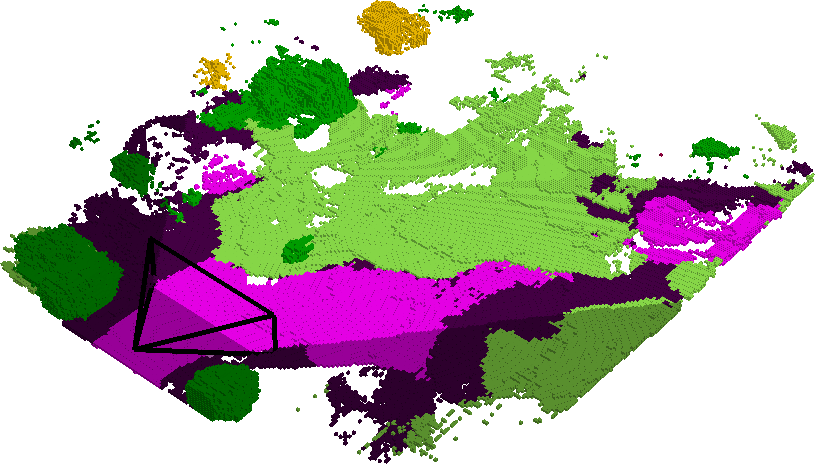} & 
			\includegraphics[width=\linewidth]{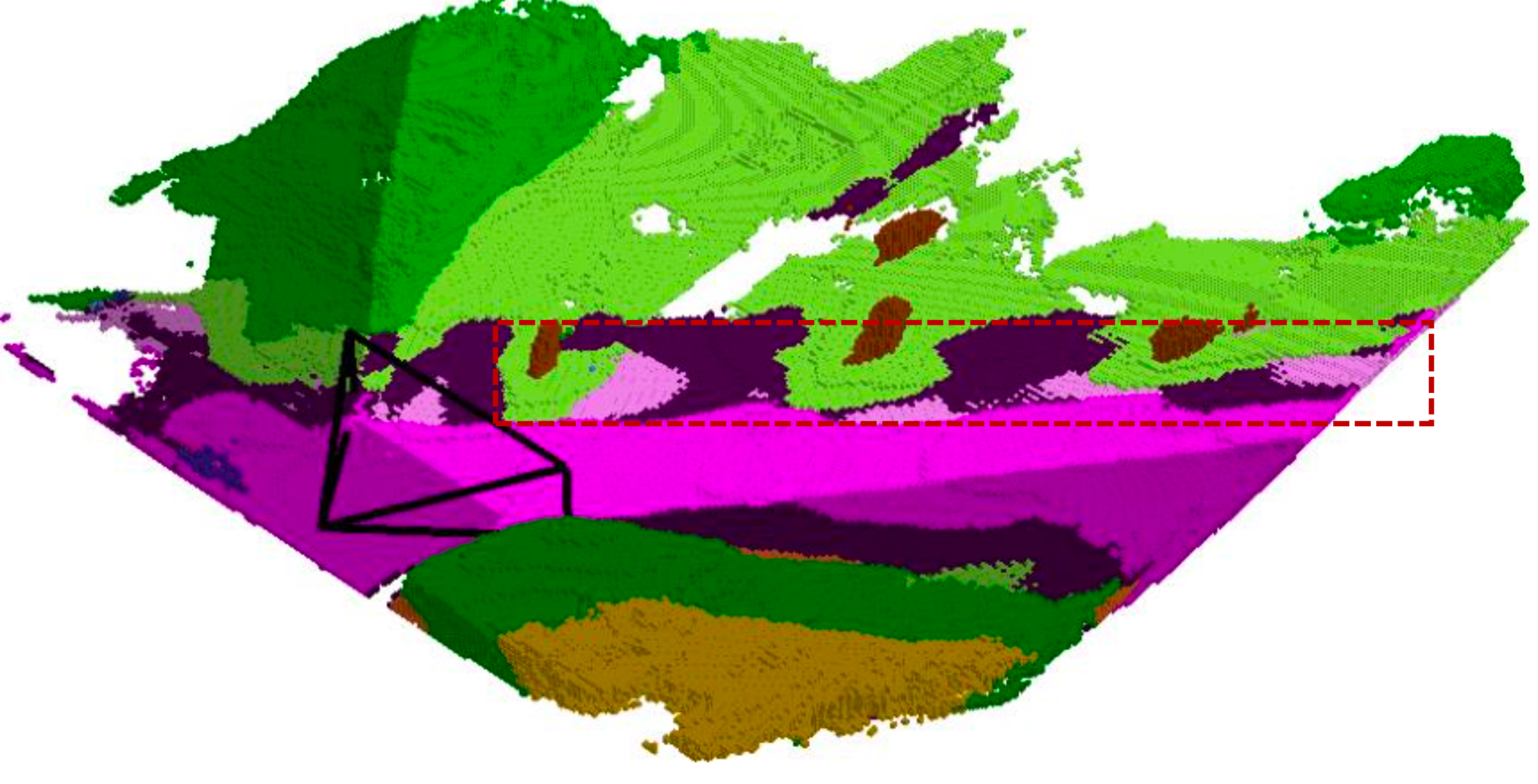} & 
			\includegraphics[width=\linewidth]
            {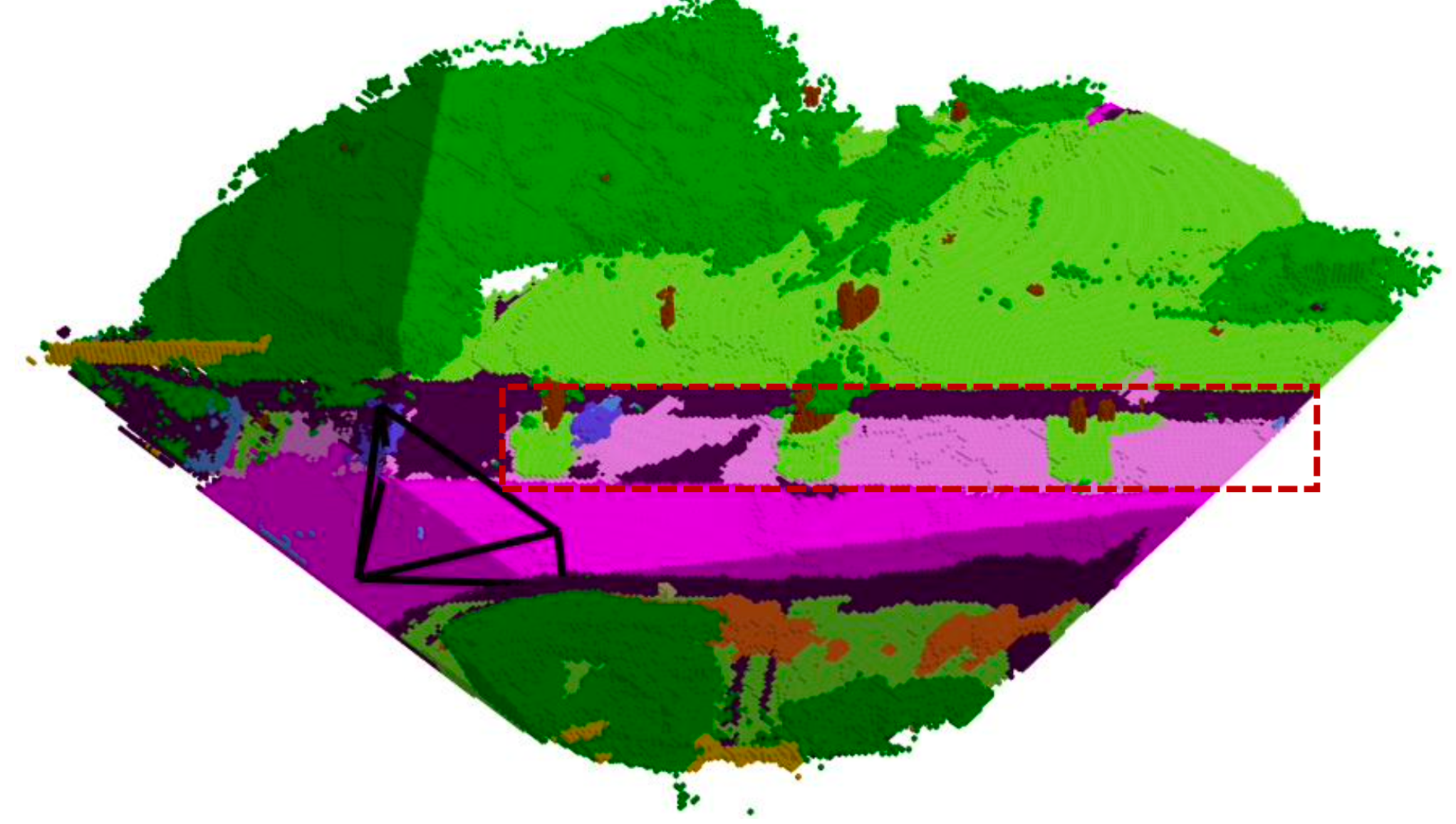} & 
			\includegraphics[width=\linewidth]
            {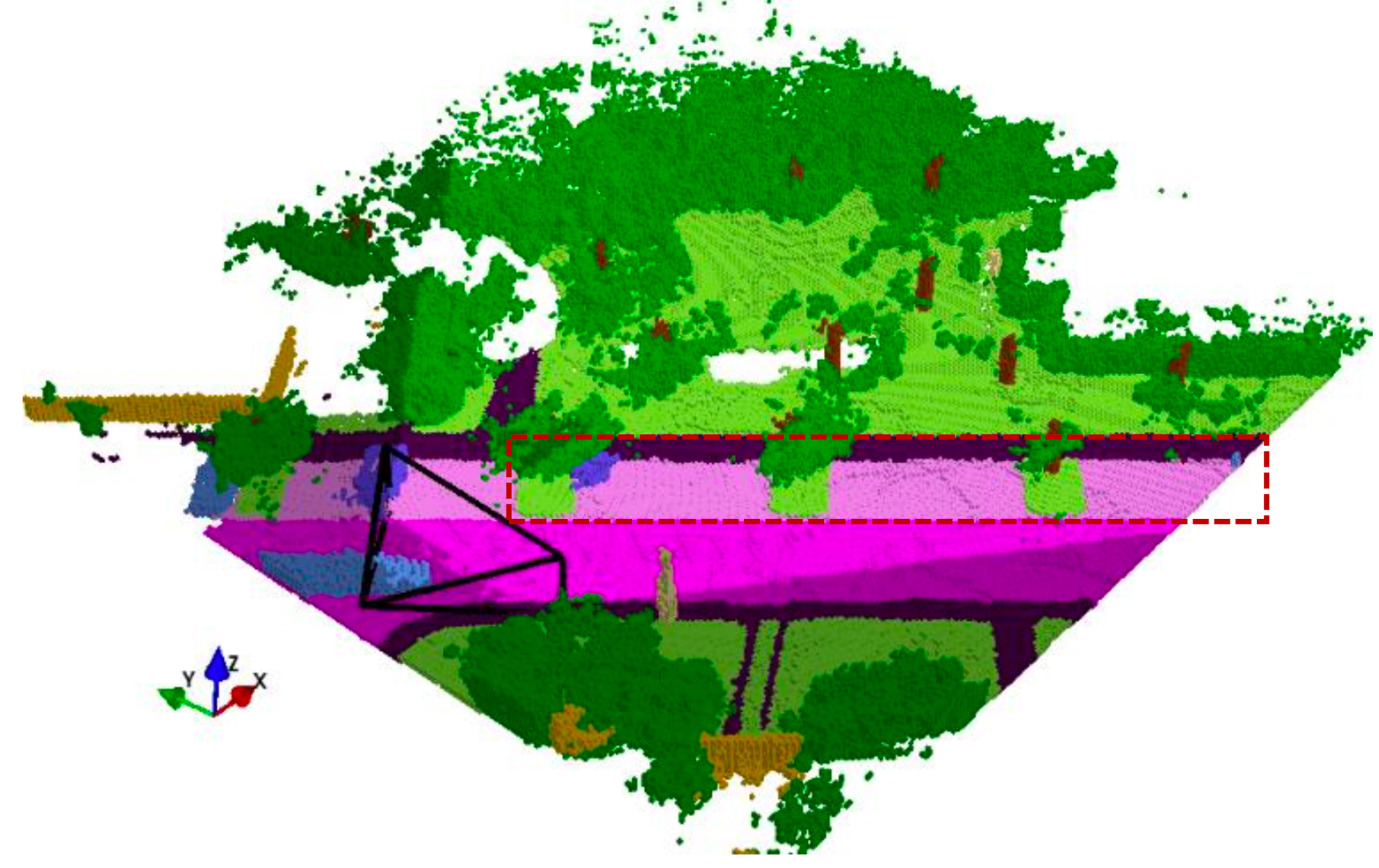} 
			\\[-0.6em]
			\includegraphics[width=\linewidth]{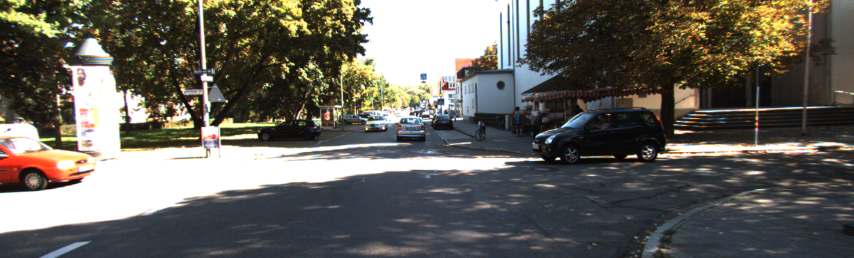} & 
			\includegraphics[width=\linewidth]{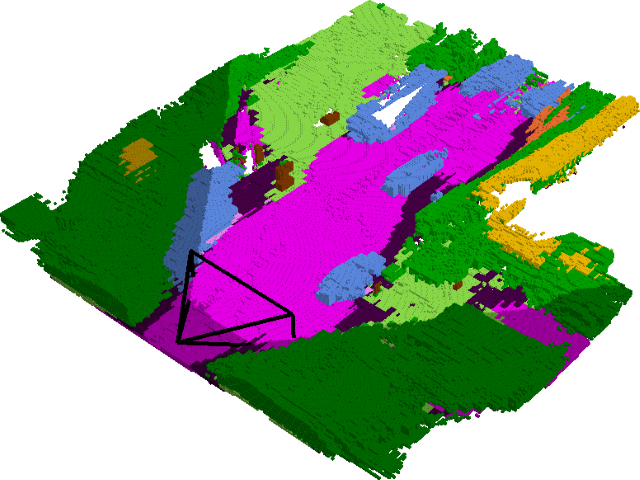} & 
			\includegraphics[width=\linewidth]{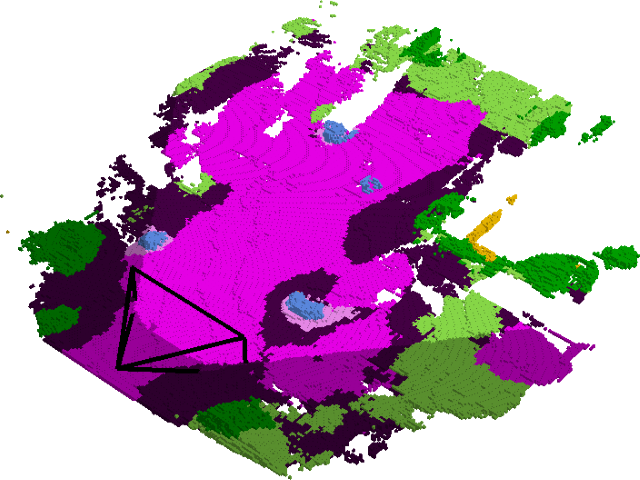} & 
			\includegraphics[width=\linewidth]{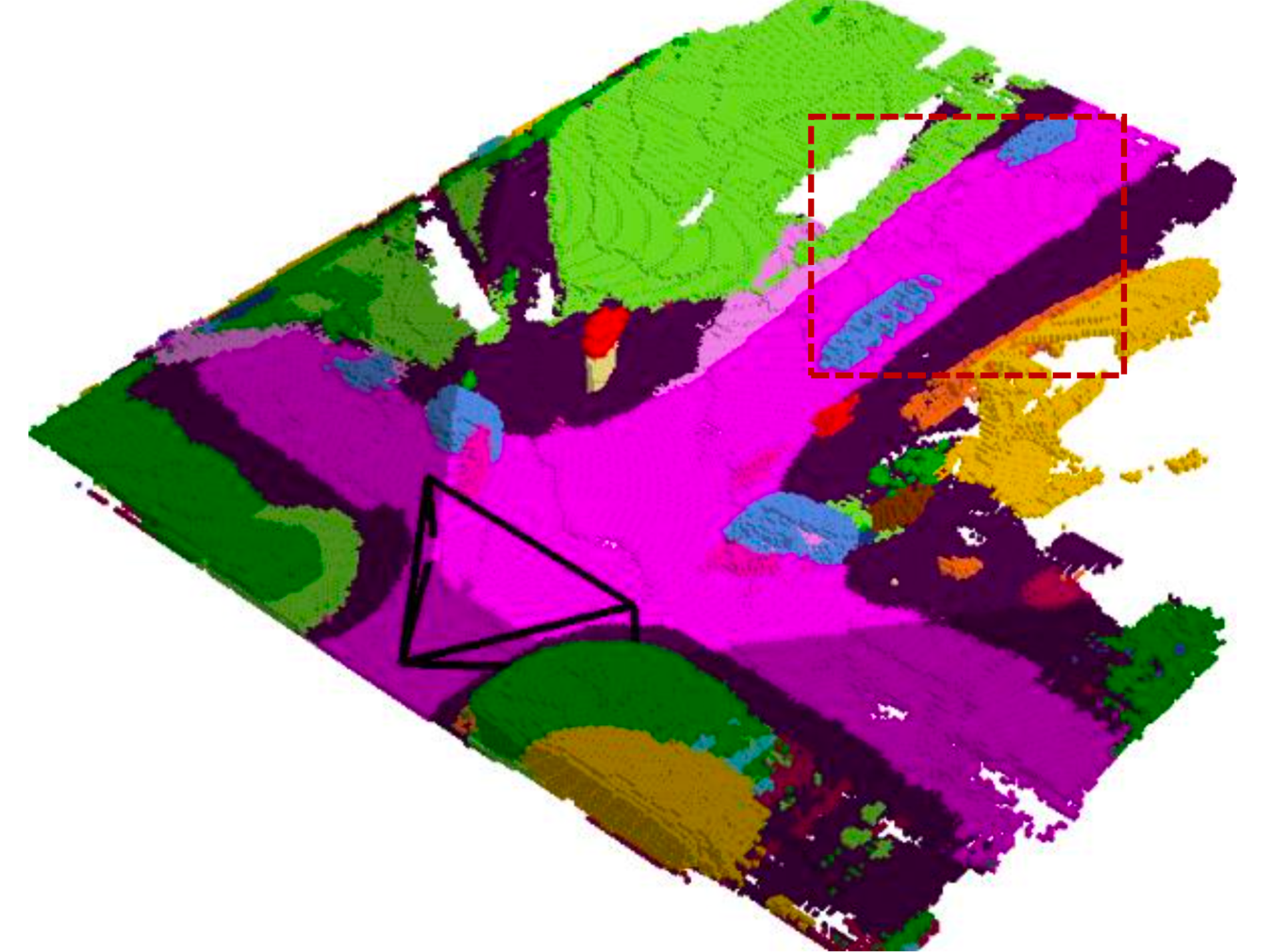} & 
			\includegraphics[width=\linewidth]
            {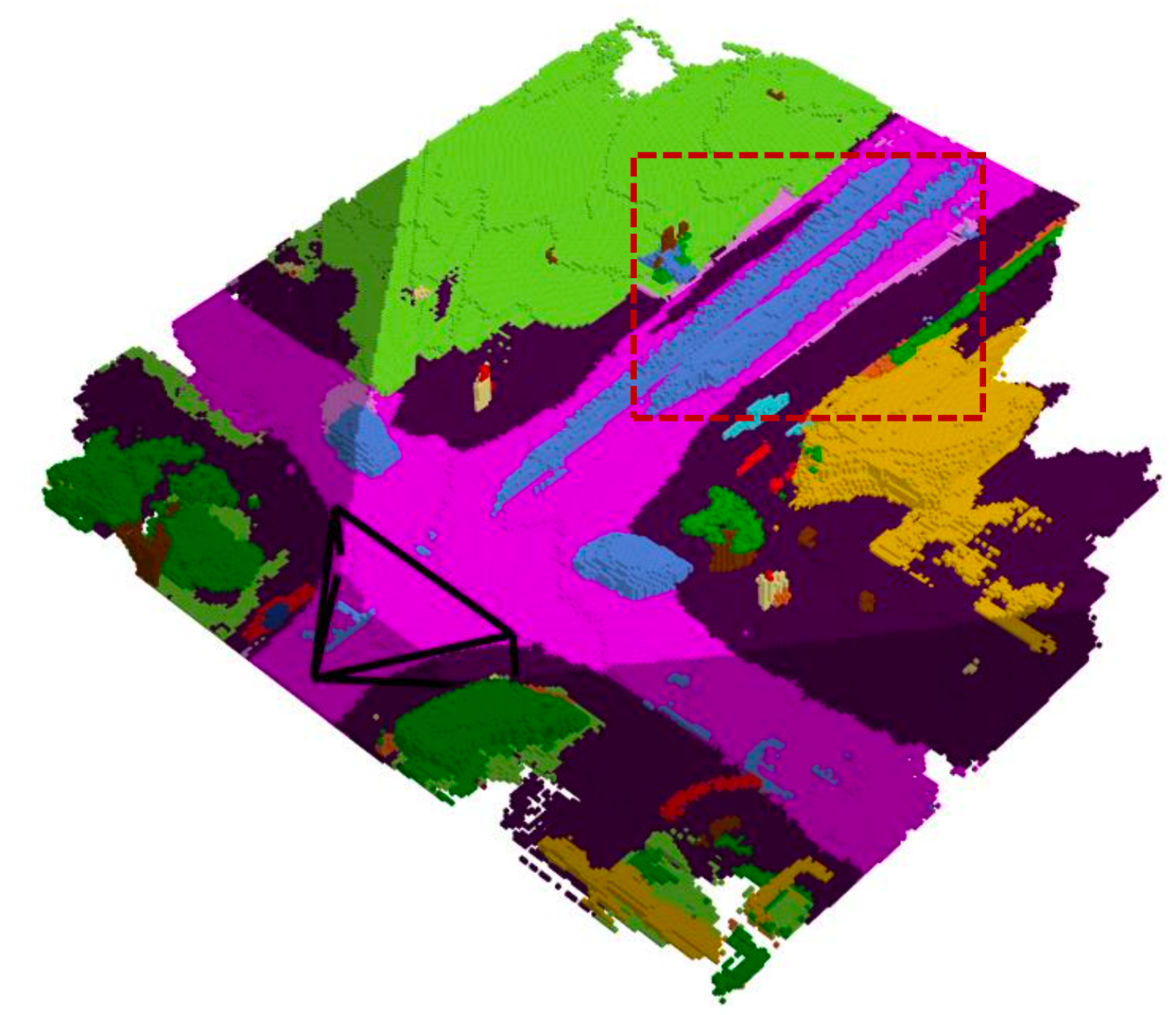} & 
			\includegraphics[width=\linewidth]
            {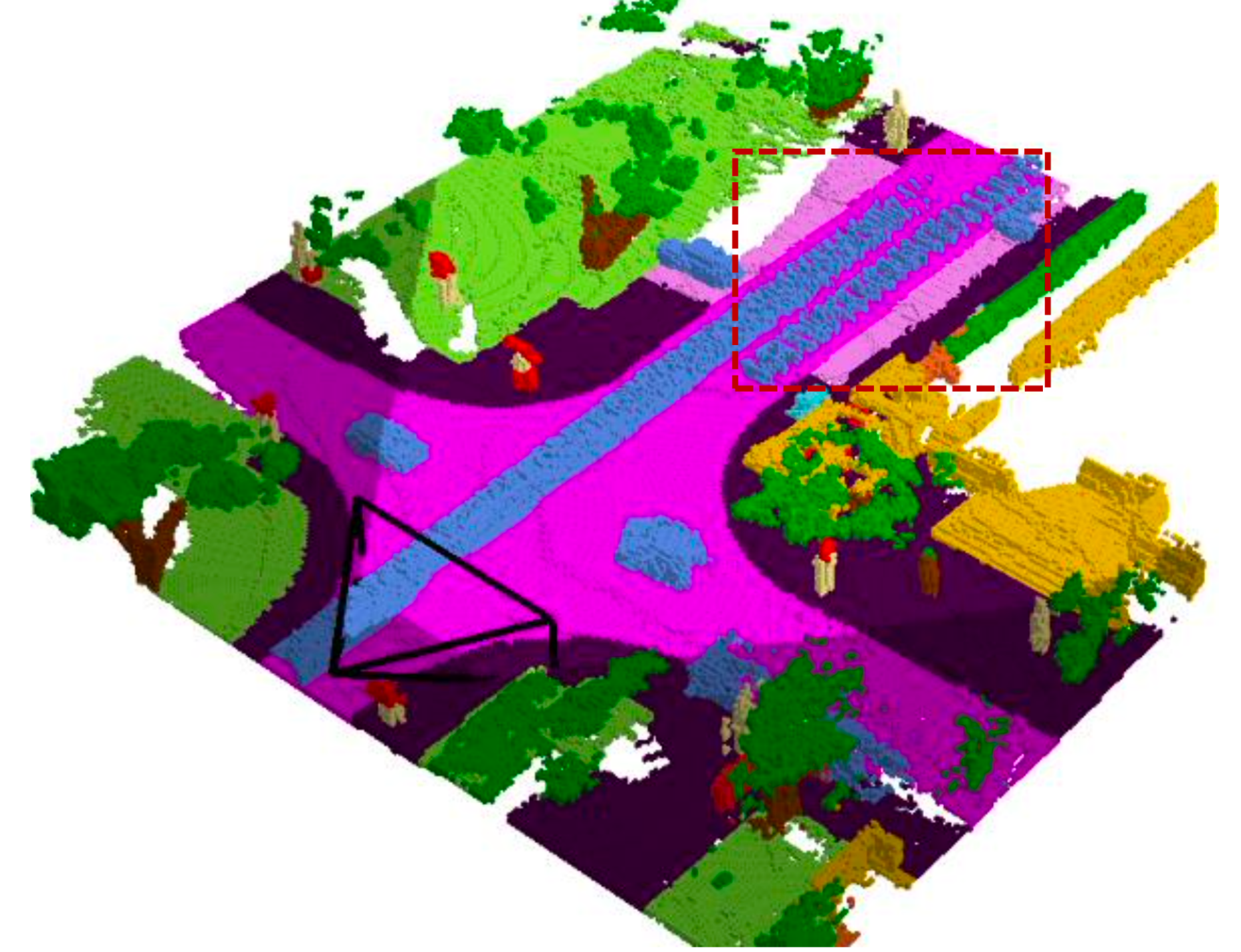} 
			\\[-0.6em]
			\includegraphics[width=\linewidth]{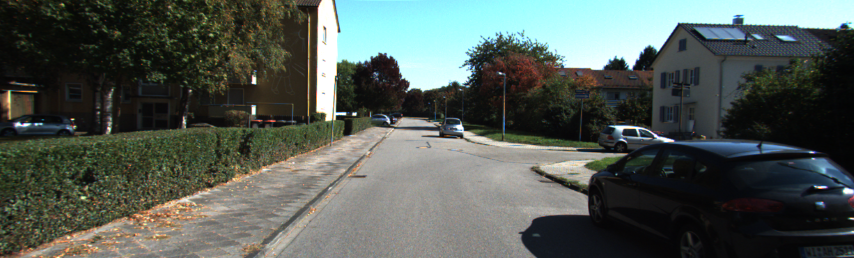} & 
			\includegraphics[width=\linewidth]
            {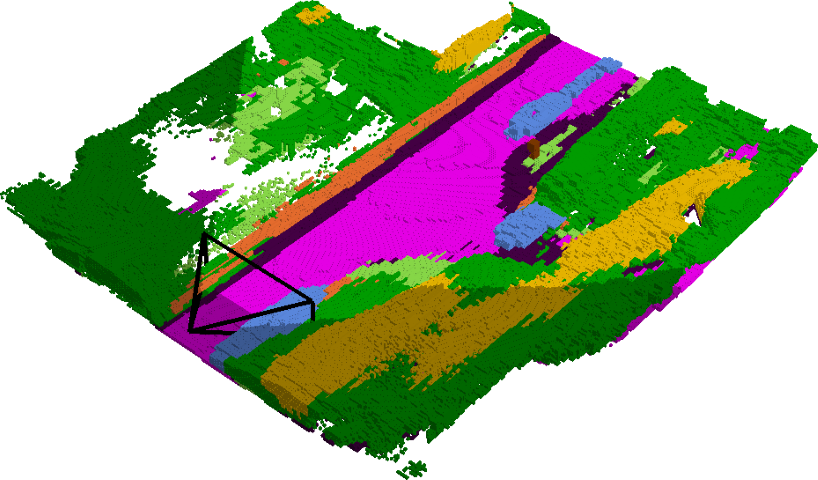} & 
			\includegraphics[width=\linewidth]{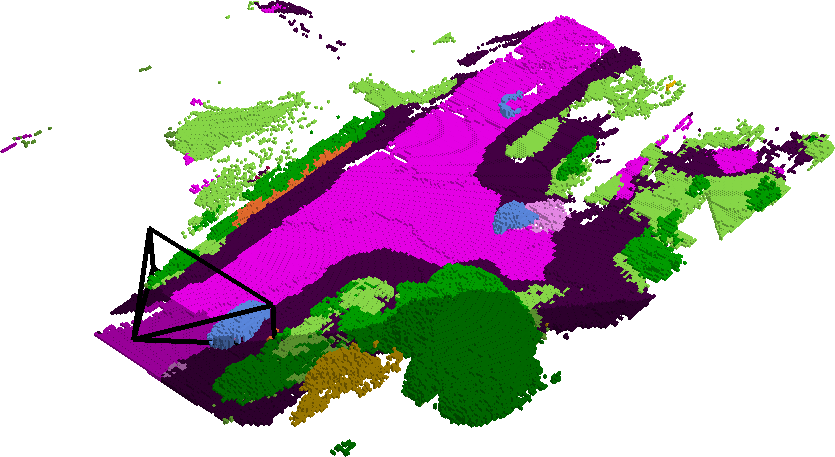} & 
			\includegraphics[width=\linewidth]{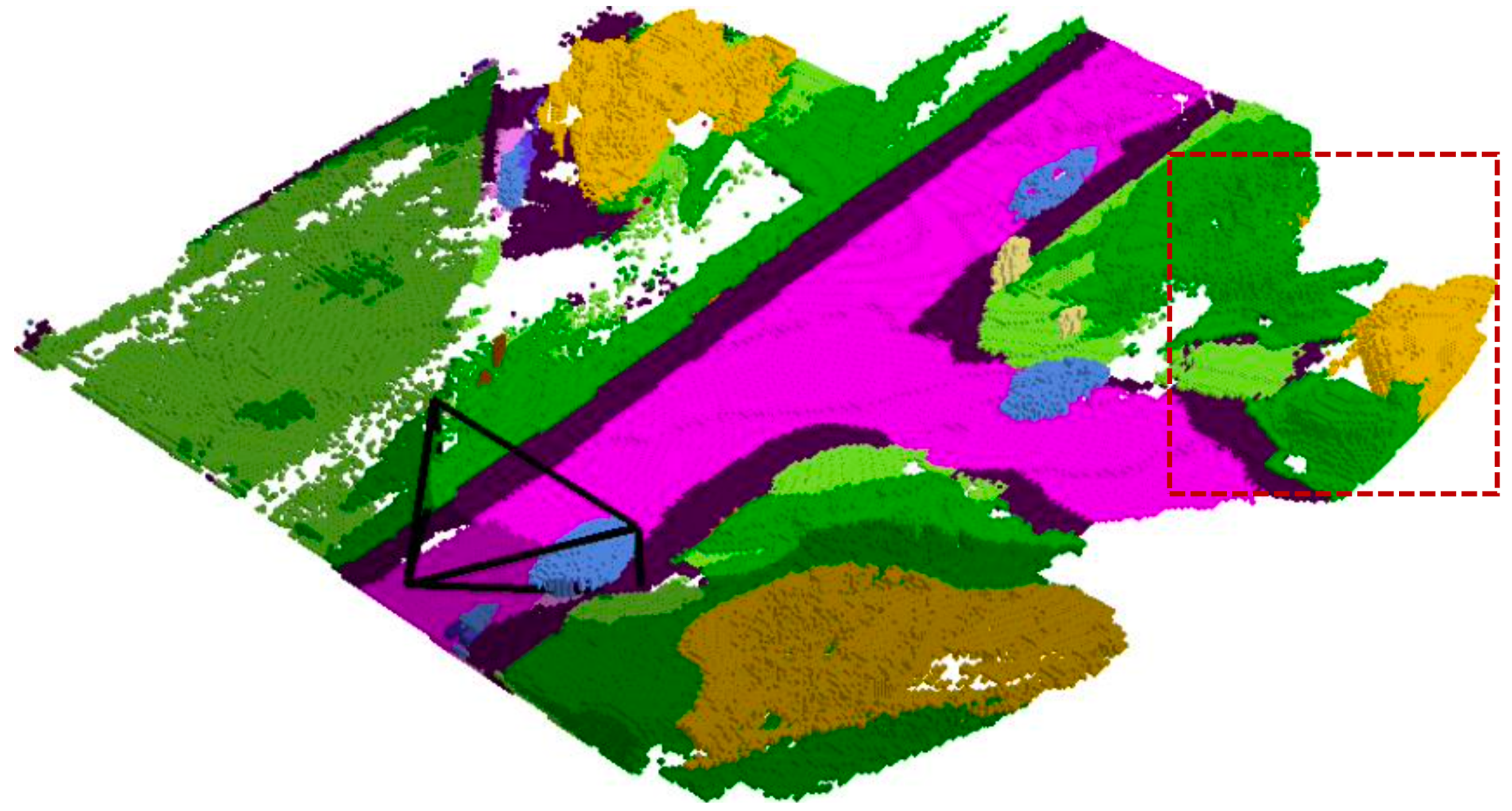} & 
			\includegraphics[width=\linewidth]
            {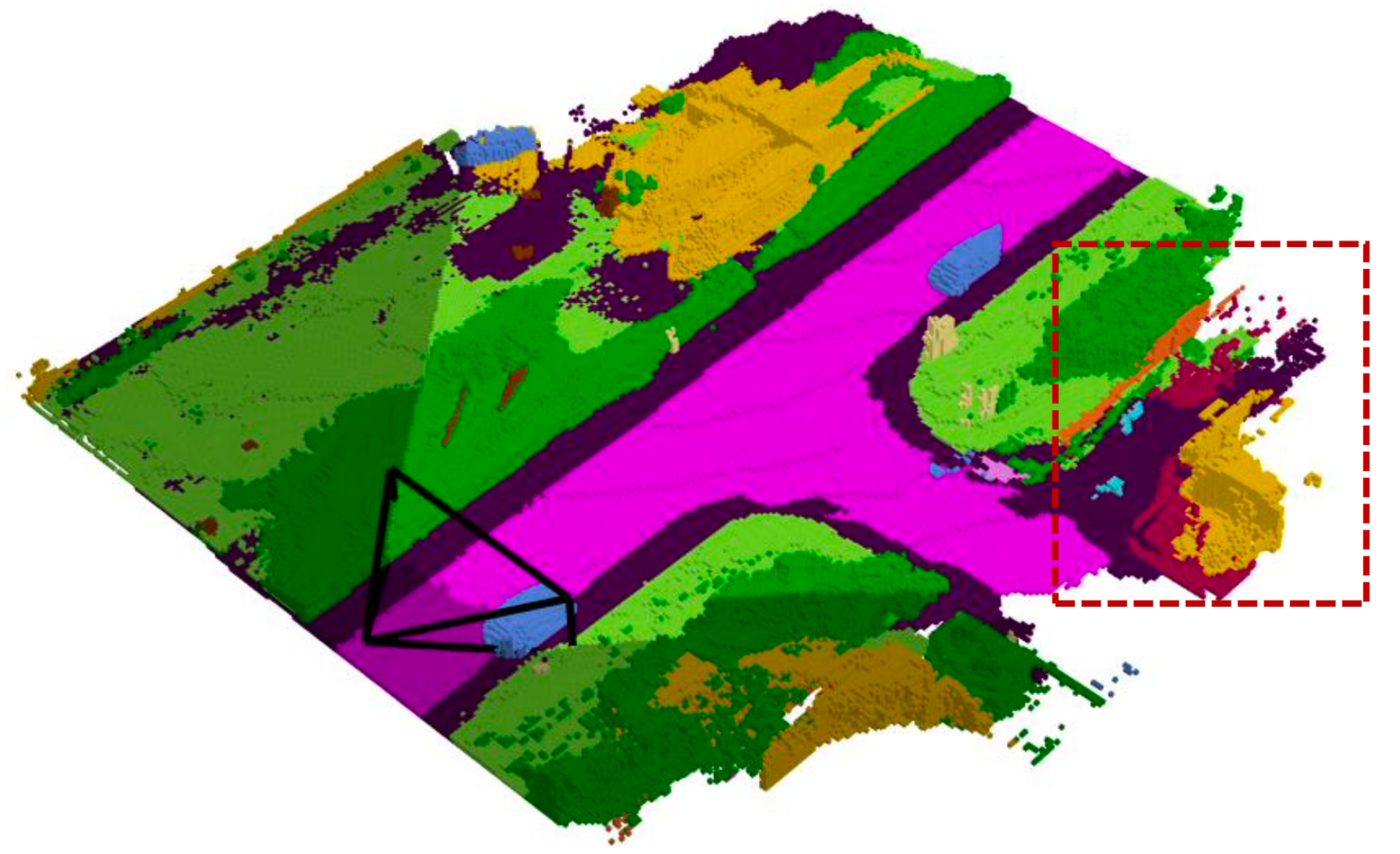} & 
			\includegraphics[width=\linewidth]
            {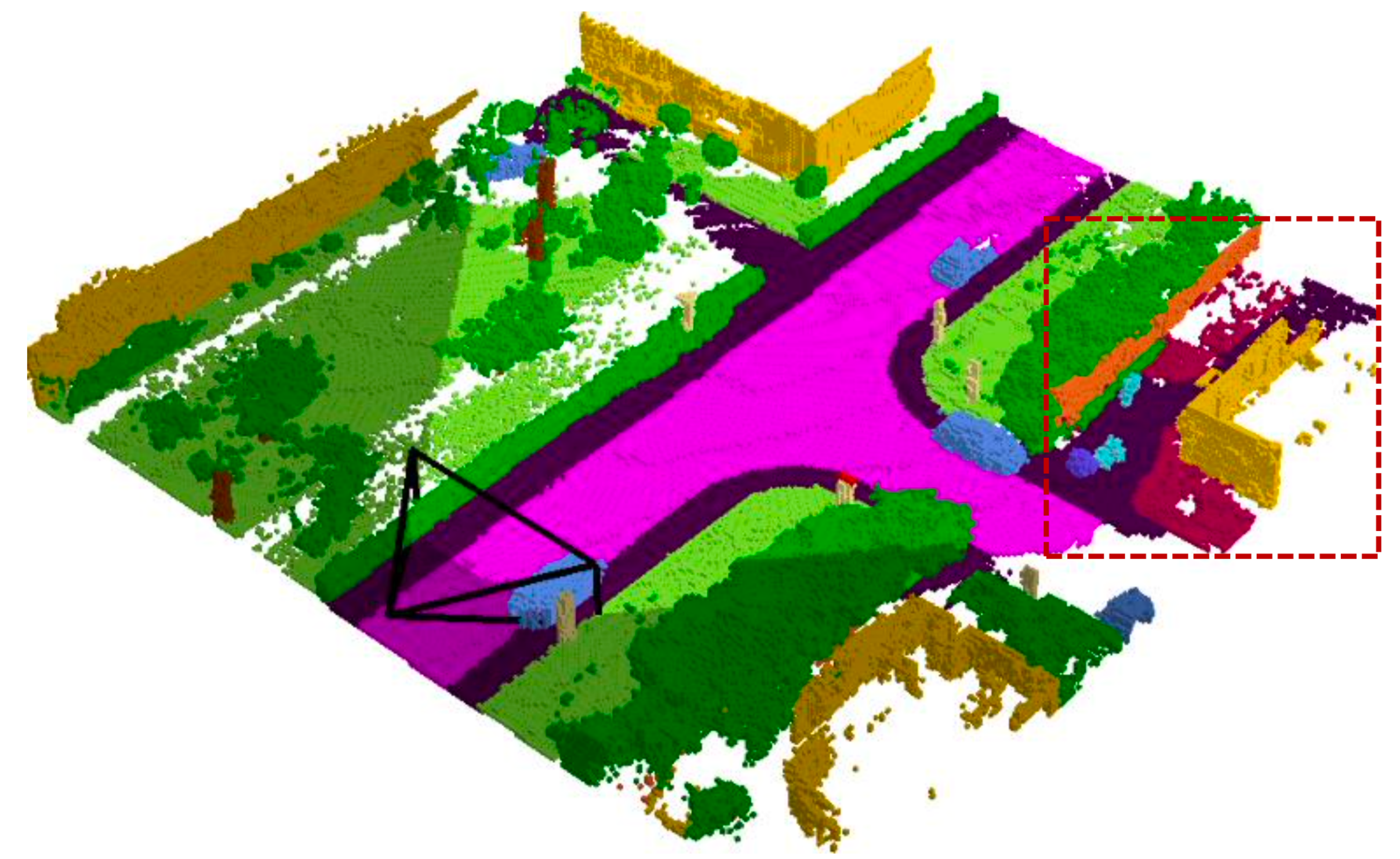} 
			\\		
			\multicolumn{6}{c}{
				\scriptsize
				\textcolor{bicycle}{$\blacksquare$}bicycle~
				\textcolor{car}{$\blacksquare$}car~
				\textcolor{motorcycle}{$\blacksquare$}motorcycle~
				\textcolor{truck}{$\blacksquare$}truck~
				\textcolor{other-vehicle}{$\blacksquare$}other vehicle~
				\textcolor{person}{$\blacksquare$}person~
				\textcolor{bicyclist}{$\blacksquare$}bicyclist~
				\textcolor{motorcyclist}{$\blacksquare$}motorcyclist~
				\textcolor{road}{$\blacksquare$}road~
				\textcolor{parking}{$\blacksquare$}parking~}\\
			\multicolumn{6}{c}{
				\scriptsize
				\textcolor{sidewalk}{$\blacksquare$}sidewalk~
				\textcolor{other-ground}{$\blacksquare$}other ground~
				\textcolor{building}{$\blacksquare$}building~
				\textcolor{fence}{$\blacksquare$}fence~
				\textcolor{vegetation}{$\blacksquare$}vegetation~
				\textcolor{trunk}{$\blacksquare$}trunk~
				\textcolor{terrain}{$\blacksquare$}terrain~
				\textcolor{pole}{$\blacksquare$}pole~
				\textcolor{traffic-sign}{$\blacksquare$}traffic sign			
			}
		\end{tabular}

        \captionsetup{font = footnotesize}
	\caption{\textbf{Qualitative results on SemanticKITTI~\cite{behley2019semantickitti} (val set).} From left to right: (a) RGB input, (b) results of AICNet$^\text{rgb}$~\cite{li2020anisotropic}, (c) results of LMSCNet$^\text{rgb}$~\cite{roldao2020lmscnet}, (d) results of  MonoScene~\cite{cao2022monoscene}, (e) ours results. NDC-Scene achieve higher voxel-level accuracy and better semantic predictions on SemanticKITTI (val set) compared with existing SSC baselines.}
 \label{fig:qualitative_kitti}
\end{figure*}
From this view of point, we propose depth adaptive attention to facilitate 3D features $\boldsymbol{X}_{i,j,k}^{T}$ in each depth-of-field to flexibly decide, which channel group of the 2D feature $\boldsymbol{X} _{i,j}^{2D}$ in the position $\boldsymbol{p}_{i,j,}^{2D}$ it is projected on, is most helpful to restore a robust representation of its depth-of-field. Formally, we divide the 2D feature $\boldsymbol{X} _{i,j}^{2D}$ with $C_{in}$ channels to $G$ groups, each with $C_{in}/G$ channels. The depth adaptive attention of $\boldsymbol{X}_{i,j,k}^{T}$ on $\boldsymbol{X}_{i,j}^{2D}$ is represented as follows:
\begin{align}
    \boldsymbol{A}_{i,j,k} = \textit{Softmax}\left(\boldsymbol{a_{i,j,k}}\right), \\
    \boldsymbol{a}_{i,j,k}^g = \left(W^Q\boldsymbol{X} _{i,j,k}^{T}\right)^\prime \cdot W^K\boldsymbol{X}_{i,j,g}^{2D}, \\
    \boldsymbol{X}_{i,j,k}^{T} = \boldsymbol{A}_{i,j,k}\boldsymbol{X} _{i,j}^{2D}.
\end{align}
$W^Q$ and $W^K$ are two projection matrices to calculate the similarity vector $\boldsymbol{a}$. This design is mostly inspired by the attention module in ~\cite{chowdhery2022palm}, which omits the value projection $W^V$ to save the computational cost. The detail structure is shown in Fig. \ref{fig:daa}.

\section{Experiment}
\label{sec:experiment}

We evaluate NDC-Scene on the didactic real-world indoor dataset NYUv2~\cite{silberman2012indoor} and outdoor SemanticKITTI~\cite{behley2019semantickitti}. We compare NDC-Scene with both state-of-the-art monocular SSC baselines and several adapted SSC baselines which requires additional depth information. Both datasets contains labeled ground truth in the target space and camera extrinsics.

\begin{table*}
	\centering
	\setlength{\tabcolsep}{0.010\linewidth}
	\centering
	\footnotesize
        \captionsetup{font=footnotesize}
	\begin{tabular}{l|ccc|cc|cc}
		\toprule
		& & & &\multicolumn{2}{c}{NYUv2} & \multicolumn{2}{c}{SemanticKITTI} \\
		Methods & w/o FA & w/o CI & w/ DA & IoU $\uparrow$ & mIoU $\uparrow$ & IoU $\uparrow$ & mIoU $\uparrow$ \\
		\midrule
		Ours & \checkmark  & \checkmark & \checkmark & \textbf{44.17} & \textbf{29.03}  & \textbf{37.24}  & \textbf{12.70} 	
		\\
        
		NDC-FA & \scalebox{0.85}[1]{$\times$}  & \checkmark & \checkmark& 42.96 \textcolor{blue}{(-1.21)} & 27.69 \textcolor{blue}{(-1.34)} & 37.15 \textcolor{blue}{(-0.09)} & 12.03 \textcolor{blue}{(-0.67)}
		\\
		NDC-CI & \checkmark  & \scalebox{0.85}[1]{$\times$} & \checkmark & 43.72 \textcolor{blue}{(-0.45)}  & 28.10 \textcolor{blue}{(-0.93)} & 37.20 \textcolor{blue}{(-0.04)} & 12.26 \textcolor{blue}{(-0.44)}
		\\
		NDC-NF & \checkmark  & \checkmark & \scalebox{0.85}[1]{$\times$} & 43.52 \textcolor{blue}{(-0.65)} & 28.22 (\textcolor{blue}{-0.81}) & 37.18 \textcolor{blue}{(-0.06)} & 11.22 \textcolor{blue}{(-0.48)}	
  		\\
		MonoScene & \scalebox{0.85}[1]{$\times$} & \scalebox{0.85}[1]{$\times$} & \scalebox{0.85}[1]{$\times$}& 42.51 \textcolor{blue}{(-1.66)} & 26.94 (\textcolor{blue}{-2.09}) & 37.12 \textcolor{blue}{(-0.12)} & 11.50 \textcolor{blue}{(-1.20)}	
		\\
		
		\bottomrule
	\end{tabular}
 
	\caption{\textbf{Components ablation.} All of our components boost performance consistently on NYUv2~\cite{silberman2012indoor} and SemanticKITTI~\cite{behley2019semantickitti}.}
	\label{tab:ablation}
\end{table*}

\textbf{NYUv2 Dataset}~\cite{silberman2012indoor} consists of 1449 scenes captured using the Kinect camera, which are represented as $240\times144\times240$ voxel grids annotated with 13 classes (11 semantic, one free, and one unknown). The shape of the RGB images is $640 \times 480$. Following previous works~\cite{cao2022monoscene}, we split the dataset with 795 training samples and 654 test samples.

\textbf{SemanticKITTI Dataset}~\cite{behley2019semantickitti} is a large-scale outdoor 3D scene semantic complement dataset, which contains Lidar scans represented as $256\times256\times32$ grids of 0.2m voxels. The dataset includes 21 classes, including 19 semantic labels, one free label, and one unknown label. We utilize the RGB images of cam-2 with a resolution of $1226\times370$ and left cropped to $1220\times370$. And follow the official train / val splits consisting 3834 and 815 samples, respectively. Following~\cite{cao2022monoscene}, we evaluate our models at full scale (\ie 1:1). Our main results and ablations are performed using the offline validation set for convenience, and the results on the hidden test set on the online server are also provided in the supplementary materials.

\paragraph{Metrics}~For the scene completion (SC) metric, we present the intersection over union (IoU) of occupied voxels regardless of their semantic classification following common practice.  For the semantic scene completion (SSC) metric, we report the mean IoU (mIoU) across all semantic categories. Note that the training and evaluation processes differ for indoor and outdoor settings because of the different depth and sparsity of the LiDAR data. To account for both scenarios, we use the more challenging evaluation metrics for all voxels, and we follow \cite{cao2022monoscene} to report the baseline results under the consistent metric.

\paragraph{Implementation Details}
Our experiments are conducted on 2 NVIDIA Tesla V100 SXM2 GPUs. Specifically, we exploit DDR module~\cite{li2020anisotropic} and 3D deconvolution layer as the 3D computation unit and the upsample operation in DDAD. As for the 3D UNet, the downsample and upsample operation are instantiated as 3D convolution and deconvolution layer, both with stride 2. The initial 2D/3D feature maps in DDAD are 15x20/15x20x16 and 39x12/39x12x32, respectively for NYUv2 and SemanticKITTI, while the output 3D feature maps of DADD are 60x80x64 and 156x48x128. We follow the loss functions in ~\cite{cao2022monoscene}. The group number $g$ of DAA is 8.

\subsection{Performance}

We compare our proposed NDC-Scene with existing strong SSC baselines designed for indoor (3DSketch~\cite{chen20203d}, and AICNet~\cite{li2020anisotropic}) or outdoor (LMSCNet~\cite{roldao2020lmscnet}) scenarios. We also compare NDC-Scene with MonoScene~\cite{cao2022monoscene} , which is the best RGB-only SSC method. Note that for the methods with more than RGB inputs, we follow \cite{cao2022monoscene} to adapt their results to RGB-only inputs. 

\paragraph{NYUv2}~Tab.~\ref{tab:NYUv2} presents the performance of NDC-Scene on NYUv2~\cite{silberman2012indoor} compared with other SSC methods, which outperforms all other methods by a considerable margin. Compared with previous state-of-the-art MonoScene~\cite{cao2022monoscene}, our method obtains better results not only on both IoU and mIoU but also on all categories, demonstrating the effectiveness and robustness of our proposed architecture. 

\paragraph{SemanticKITTI}~We compare the results under the outdoor scenarios on SemanticKITTI~\cite{behley2019semantickitti} in Tab.~\ref{tab:SemanticKITTI}. For all categories including small (\eg\ car, bicycle, and person) and large(\eg\ building, truck, and road) semantics, our method beat the baselines significantly, showing that our method can adapt to different scenarios.

\paragraph{Qualitative performance}~In Fig.~\ref{fig:qualitative_NYU}, we present visualization results to qualitatively evaluate the effectiveness of our NDC-Scene on NYUv2~\cite{silberman2012indoor} dataset. We can see that the proposed NDC-Scene has the ability to handle a wide range of objects with diverse shapes, resulting in more precise scene layout and instance-level information than other SSC baselines. Fig.~\ref{fig:qualitative_kitti} illustrates the qualitative results on SemanticKITTI~\cite{behley2019semantickitti}. Our NDC-Scene demonstrates satisfactory performance in discerning the accurate depth range of objects like terrains, trunks and cars, as highlighted in the red boxes. This improvement holds great significance in large-scale outdoor scenarios and serves as compelling evidence that we have greatly relieved the issue of Feature Ambiguity exposed by MonoScene~\cite{cao2022monoscene}.

\subsection{Ablation Study}

\begin{figure}[!t]
    \centering
    
    \captionsetup{font=footnotesize}
    \includegraphics[width=0.45\textwidth]{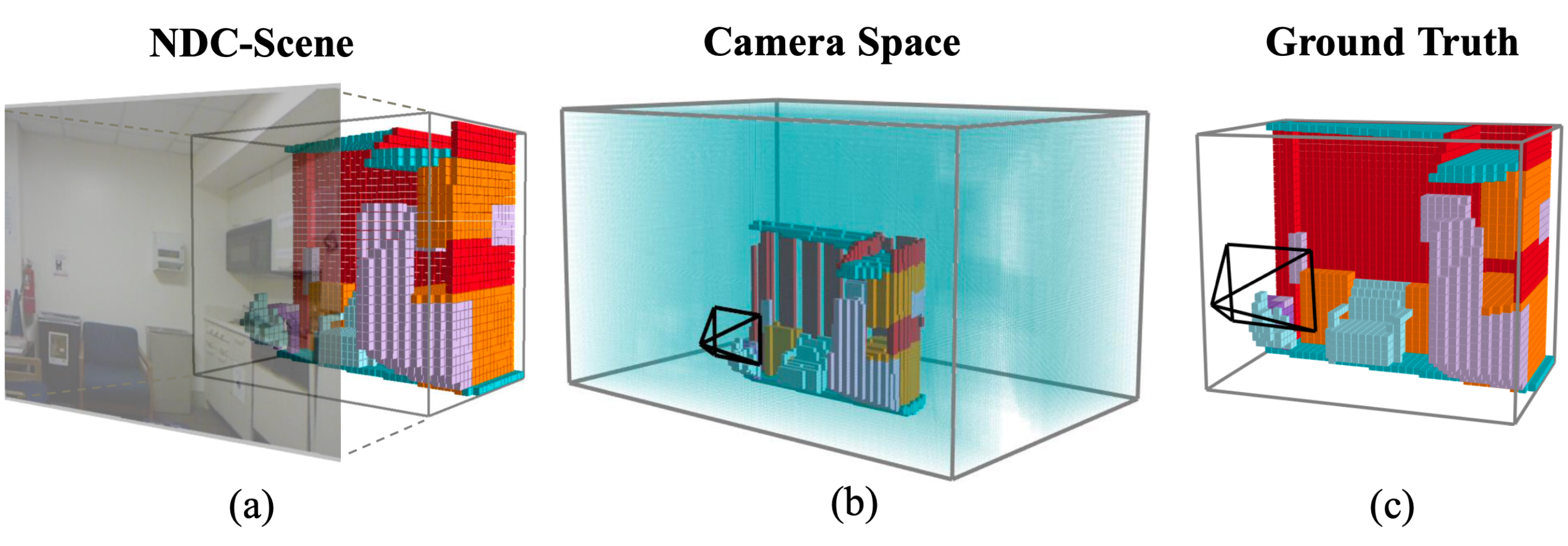}
   
    \caption{\textbf{Qualitative comparison} of the normalized device coordinates space and the camera space. In (b), the voxels of $S^R$ that are close to the camera are mostly out of the image when projected on the camera plane. Thus the in-image voxels are quite sparse. In contrast, in (a) the voxels with different depths are all uniformly projected on the camera plane.}
    
    \label{fig:imbalance_vis}
\end{figure}

We design comprehensive experiments to verify the capacity of the proposed method in solving the identified problems of feature ambiguity, pose ambiguity and computation imbalance.


\paragraph{Feature Ambiguity}
To evaluate NDC-Scene's ability in relieving the feature ambiguity problem, we replace the proposed dual decoder with a heavy 3D UNet with the same amount of FLoSP, the input features of which are lifted via FLoSP. This variant loses the ability of implicitly restoring the semantics among different depths, named as NDC-FA. For the detailed architecture of NDC-FA, please refer to the supplementary. Tab.~\ref{tab:ablation} shows that, with the introduced feature ambiguity, the performance in both geometry ([-1.21, -0.09] IoU) and semantics ([-1.34, -0.67] mIoU) degrades a lot.


\paragraph{Computation Imbalance}
Also, NDC-CI is designed to verify the capacity of NDC-Scene in solving the computation imbalance problem, Where the 3D feature map in DADD is replaced with a camera space $S^R$, with coordinates $\boldsymbol{p}_{i,j,k}^{R}=\left(x_{i,j,k}^R, y_{i,j,k}^R, d_{i,j,k}^R \right)$ uniformly distributing in $S^R$. As the 2D pixels $\boldsymbol{p}_{i,j}^{2D}$ and $\boldsymbol{p}_{i,j,k}^{R}$ are also related by Eq.~\ref{eq:prospective} in the same way of $\boldsymbol{p}_{i,j}^{2D}$ and $\boldsymbol{p}_{i,j,k}^{T}$, 3D convolution in $S^R$ also suffers from the CI problem. A intuitive comparison between $S^R$ and the proposed $S^N$ is illustrated in Fig.~\ref{fig:imbalance_vis}. We notice that NDC-CI degrades much more in semantics ([-0.93, -0.44] mIoU) than in geometry ([-0.45, -0.04] IoU), as shown in Tab.~\ref{tab:ablation}.

\paragraph{Depth Adaptive Attention} 
We compare DAA with a naive fusion approach which direct adds the 2D feature on $\boldsymbol{p}_{i,j}^{2D}$ to all 3D features that projects at $\boldsymbol{p}_{i,j}^{2D}$, named as NDC-NF. Accoring to Tab.~\ref{tab:ablation}, DAA compensates for ~30\% of the performance gain compared to MonoScene.
\begin{table}
	\centering
        \captionsetup{font=footnotesize}
	\setlength{\tabcolsep}{0.010\linewidth}
	\centering
	\footnotesize
	\begin{tabular}{l|cc|cc}
		\toprule
		& \multicolumn{2}{c}{Ours(NYUv2)} & \multicolumn{2}{c}{MonoScene(NYUv2)} \\
		& IoU $\uparrow$ & mIoU $\uparrow$ & IoU $\uparrow$ & mIoU $\uparrow$ \\
		\midrule
			
	$\theta$ = $0^\circ$ &  \textbf{44.17} & \textbf{29.03}  & \textbf{42.51}  & \textbf{26.94} 	
		\\
		
		$\theta$ = $5^\circ$ &  42.88 \textcolor{blue}{(-1.29)} & 28.28 \textcolor{blue}{(-0.75)} & 38.99 \textcolor{blue}{(-3.52)} & 23.20 \textcolor{blue}{(-3.74)}
		\\
		$\theta$ = $10^\circ$ & 39.07 \textcolor{blue}{(-5.10)}  & 24.64 \textcolor{blue}{(-4.39)} & 33.52 \textcolor{blue}{(-8.99)} & 20.15 \textcolor{blue}{(-6.79)}
		\\
		$\theta$ = $15^\circ$ & 36.74 \textcolor{blue}{(-7.43)} & 22.39 (\textcolor{blue}{-6.64}) &  30.05\textcolor{blue}{(-12.46)} & 16.71 \textcolor{blue}{(-10.23)}	
		\\
		
		\bottomrule
	\end{tabular}
	\caption{\textbf{Ablation study} for the robustness to pose ambiguity on NYUv2~\cite{silberman2012indoor}.}
    
	\label{tab:ablation_pa}
\end{table}
\paragraph{Pose Ambiguity}
For PA, we randomly rotate target space as well as the SSC label with an angle uniformly sampled in $\left[0, \theta\right]$, with $\theta$ in $\left[5^\circ, 10^\circ, 15^\circ\right]$, and compare the performance degradation of NDC-Scene and MonoScene to validate the robustness of NDC-Scene in the choice of extrinsic camera parameters. As revealed in Tab.~\ref{tab:ablation_pa}, the performance of NDC-Scene degrades much slower than MonoScene as the increase of $\theta$, which verifies the strong ability of NDC-Scene in handling pose ambiguity.

\section{Conclusion}

To conclude, our study comprehensively explores the critical challenges encountered by the present state-of-the-art techniques in monocular 3D semantic scene completion. To overcome these challenges, the proposed method introduces a novel Normalized Device Coordinates (NDC) space predictor technique, which effectively extends the 2D feature map to a 3D space by progressively restoring the dimension of depth with deconvolution operations. By transferring the majority of computation from the target 3D space to the proposed normalized device coordinates space, the proposed approach leads to enhanced performance in monocular SSC tasks. Furthermore, the study proposes the use of a Depth-Adaptive Dual Decoder, which facilitates the simultaneous upsampling and fusion of the 2D and 3D feature maps, thereby improving overall performance.
\section{Acknowledgements}
This project is funded in part by National Key R\&D Program of China Project 2022ZD0161100, by the Centre for Perceptual and Interactive Intelligence (CPII) Ltd under the Innovation and Technology Commission (ITC)’s InnoHK, by General Research Fund of Hong Kong RGC Project 14204021. Hongsheng Li is a PI of CPII under the InnoHK.

\newpage
{\small
\bibliographystyle{ieee_fullname}
\bibliography{egbib}
}

\end{document}